\documentclass{ieeeaccess}
\usepackage{cite}
\usepackage{amsmath,amssymb,amsfonts}
\usepackage{algorithmic}
\usepackage{graphicx}
\usepackage{textcomp}
\usepackage{booktabs}
\usepackage{multirow}
\usepackage{multicol}
\usepackage{amsmath}
\usepackage{commath}

\def\BibTeX{{\rm B\kern-.05em{\sc i\kern-.025em b}\kern-.08em
    T\kern-.1667em\lower.7ex\hbox{E}\kern-.125emX}}

\begin{document}
\history{Date of publication xxxx 00, 0000, date of current version xxxx 00, 0000.}
\doi{10.1109/ACCESS.2023.0322000}

\title{Memory-Efficient Continual Learning Object Segmentation for Long Videos}
\author{\uppercase{Amir Nazemi}\authorrefmark{1},
\uppercase{Mohammad Javad Shafiee}\authorrefmark{1}, \uppercase{Zahra Gharaee} \authorrefmark{1}, and \uppercase{Paul Fieguth} \authorrefmark{1}}

\address[1]{Vision \& Image Processing Lab, Department of Systems Design Engineering,University of Waterloo,\\ Waterloo, Ontario, Canada~(e-mail: amir.nazemi, mjshafiee, zahra.gharaee, paul.fieguth@uwaterloo.ca)}
\tfootnote{We thank Microsoft Office Media Group, NSERC Alliance, and the Digital Research Alliance of Canada (alliancecan.ca) for their generous support of this research.}

\markboth
{Nazemi \headeretal: Preparation of Papers for IEEE TRANSACTIONS and JOURNALS}
{Nazemi \headeretal: Preparation of Papers for IEEE TRANSACTIONS and JOURNALS}

\corresp{Corresponding author: Amir Nazemi (e-mail: amir.nazemi@uwaterloo.ca).}

\begin{abstract}
Recent state-of-the-art semi-supervised Video Object Segmentation (VOS) methods have shown significant improvements in target object segmentation accuracy when information from preceding frames is used in segmenting the current frame. In particular, such memory-based approaches can help a model to more effectively handle appearance changes (representation drift) or occlusions. Ideally, for maximum performance, Online VOS methods would need all or most of the preceding frames (or their extracted information) to be stored in memory and be used for online learning in later frames. Such a solution is not feasible for long videos, as the required memory size grows without bound, and such methods can fail when memory is limited and a target object experiences repeated representation drifts throughout a video.

We propose two novel techniques to reduce the memory requirement of Online VOS methods while improving modeling accuracy and generalization on long videos. 
Motivated by the success of continual learning techniques in preserving previously-learned knowledge, here we  propose Gated-Regularizer Continual Learning (GRCL), which improves the performance of any Online VOS subject to limited memory, and a Reconstruction-based Memory Selection Continual Learning (RMSCL), which empowers Online VOS methods to efficiently benefit from stored information in memory. We also analyze the performance of a hybrid combination of the two proposed methods.

Experimental results show that the proposed methods are able to improve the performance of Online VOS models by more than \(8\%\), with improved robustness on long-video datasets while maintaining comparable performance on short-video datasets such as DAVIS16, DAVIS17, and YouTube-VOS18.
\end{abstract}

\begin{keywords}
Video Object Segmentation, Continual Learning, regularization-based solutions, replay-based methods.\end{keywords}

\titlepgskip=-21pt

\maketitle

\section{Introduction}
\label{sec1}
Video object segmentation (VOS) aims to extract an accurate pixel-wise object mask in each frame of a given video. Broadly, proposed VOS algorithms can be divided into two different streams: i) semi-supervised or one-shot VOS, when the ground truth masks of the target objects are provided in at least one frame at inference time, and ii) unsupervised VOS, when no information about the objects is provided. The focus of this paper is on the former context, that of semi-supervised VOS.

The classic and initial solution for a semi-supervised VOS problem is to fine-tune the trained VOS model on the given information (i.e., the given object mask), separately for each test video. This ideal is not feasible, due to the limited training samples, the VOS model size, and the time-consuming fine-tuning process. In practice, online learning-based VOS approaches \cite{oh2019video,robinson2020learning,joint,bhat2020learning} address these challenges by introducing efficient training (fine-tuning) mechanisms and keeping some amount of information in a memory to augment the training set for model fine-tuning. 

These approaches proceed on the assumption that sufficient memory is available at inference time, and that there are no limitations in storing and exploiting information. It is also assumed that an object representation is not undergoing significant shifts between frames, such that the information stored in the memory is somehow representative of the target object in the query frame.  In practice, these assumptions hold poorly, at best, and particularly in long videos it is common to experience significant representation drift of the target object. Such a drift can lead to drastic drops in performance, particularly when there is a limitation on the amount of memory available to store past object representations.   
A second bottleneck of Online VOS is its limitation to learn useful information from memory. As more training data (more frames of video) become available in the memory,  Online VOS methods have difficulty to extract and learn discriminative  information~\cite{cheng2022xmem}, due to their limited online model size and training process, since Online VOS prefers training small models on limited memory over few epochs. Clearly these issues become increasingly problematic on long video sequences, which are the focus of this paper.

We reformulate semi-supervised VOS as online continual learning~\cite{parisi2020online}, which benefits from two disjunctive solutions with a small fixed working memory to process long video sequences:
\begin{itemize}
    \item In Section~\ref{sec:cl}, a Gated-Regularizer Continual Learning (GRCL) is proposed to improve the performance of Online VOS by preserving and consolidating the acquired knowledge from the target objects in preceding frames while limiting the required memory. 
    \item A very different approach is developed in Section~\ref{sec:rms}, where we propose a Reconstruction-based Memory Selection Continual Learning (RMSCL) method which is able to augment any Online VOS framework and improves its performance, particularly on long videos.
\end{itemize}
The GRCL is inspired from prior-based continual learning~\cite{de2021continual,chen2021overcoming}, whereas the latter RMSCL is motivated by rehearsal methods in continual learning~\cite{atkinson2021pseudo,castro2018end,wu2019large,zhao2020maintaining}. We apply the proposed methods to two state-of-the-art Online VOS algorithms, LWL~\cite{bhat2020learning} and Joint~\cite{mao2021joint}, both subject to a fixed memory.  Our experimental results show an improvement of both LWL and Joint, particularly on long video sequences.

\section{Related Work}\label{sec2}
The primary objective of our work is to address online video object segmentation, specifically when dealing with long video sequences. Our objective particularly relates to the instances which are preserved in a memory for future selection and usage in the continuation of the learning process. 
We begin by overviewing baselines, state-of-the-art memory-based approaches, and methods proposed in continual learning.

We present feature selection methods with a wide range of applications in various domains such as machine learning, data mining and computer vision which can potentially be used as memory selection for VOS.
Finally, we introduce several solutions available in the literature addressing the learning challenges of long video sequences.

\subsection{Memory-based Approaches}
Memory-based approaches~\cite{seong2021hierarchical,zhou2019enhanced,oh2019video,robinson2020learning,joint,bhat2020learning,cheng2022xmem,cheng2023putting,wang2023look,zhang2023joint} try to address semi-supervised VOS by storing representations and predicted output masks of preceding frames in a memory, and then to use them for evaluating the current frame.

Within this strategy, there are different approaches to retrieve information from the dynamic model's memory. One solution is to update (fine-tune) a small model on the memory proposed by the online learning methods \cite{caelles2017one,VoigtlaenderL17,robinson2020learning,bhat2020learning,liu2020meta}. 

A second solution is to propagate the information of the most recent predicted object masks \cite{perazzi2017learning} or features representation of preceding frames\cite{hu2017maskrnn,ventura2019rvos} proposed by the recurrent methods, and a third solution is to send a query to retrieve some information of visited frames and their representation stored in the memory proposed by the query-based methods \cite{oh2019video,hu2021learning,xie2021efficient,cheng2021rethinking,yang2021associating,lin2021query,liu2022global,seong2020kernelized,seong2021hierarchical,cheng2022xmem,wang2023look}.

The approach proposed in this paper stems from the online learning methods, and will be compared to state-of-the-art query-based methods. 

\subsubsection{Query-based Methods} Among the query-based methods is STM~\cite{oh2019video}, which uses a similarity matching algorithm to retrieve encoded information from the memory and pass it through a decoder to produce an output. 

STM performs global matching between the query and memory frames; however, in VOS, a valid assumption is to consider the locality of the target object’s appearance in memory frames. Therefore, RMNet~\cite{xie2021efficient} developed a local-to-local matching algorithm that considers the local area where the target objects appeared in previous frames.

By limiting the potential correspondences between two consecutive frames to a local window and providing kernel guidance to the non-local memory matching, HMMN~\cite{seong2021hierarchical} offers kernel-based memory matching as a means of achieving temporal smoothness. HMMN uses tracking of the most likely relationship between a memory pixel and a query pixel to match distant frames.
Unlike STM, which generates a specified memory bank for each object in the video, STCN~\cite{cheng2021rethinking} constructs a model that uses an affinity matrix based on RGB relations to learn all object relations beyond just the labeled ones. An object goes through the same affinity matrix for feature transfer when querying.

LCM~\cite{hu2021learning} suggests using a memory strategy to recover pixels globally and to learn pixel position consistency for more accurate segmentation in order to deal with appearance changes and deformation.

In order to leverage the fine-grained features of instance segmentation (IS), ISVOS~\cite{wang2023look} suggest a two-branch network: the VOS branch performs spatial-temporal object level matching with the memory bank, while their proposed IS branch explores the instance details of the objects in the current frame. They include instance-specific information into the query key using well-learned object queries from the IS branch, and then perform matching.
A recent unified VOS framework, Joint-Former ~\cite{zhang2023joint}, represents the three characteristics of feature, correspondence, and dense memory. The primary structure of Joint-Former is its Joint Block, which propagates target information to current tokens and compressed memory tokens while extracting features using attention mechanism.
Cutie~\cite{cheng2023putting} benefits from a query-based object transformer that interacts with bottom-up pixel features, an object-level memory, and a small number of object queries that are continuously generated. While high-resolution feature maps are kept for exact segmentation, object queries offer an overview of the target item.

\subsubsection{Online Learning-based Methods} On the other hand, there are online learning-based methods which learn the new object appearance within an online learning-based approach~\cite{kivinen2004online,bhat2020learning,joint} simultaneously at inference time. In this scenario, instead of using a query-based (matching-based) algorithm on each frame, a small latent model network so called target model, is updated every $\Delta_\mathrm{C}$ frames which is eventually used to produce the updated information about each video frame.

The target model proposed by FRTM~\cite{robinson2020learning}, LWL~\cite{bhat2020learning} and the induction branch of JOINT~\cite{joint} is formulated as a small convolutional neural network, which performs online learning on the available training data in the memory. As such, these methods can provide an efficient yet effective dynamic update process for VOS frameworks.

While target model-based approaches improve the performance of VOS, the effectiveness of online learning algorithms is highly dependent on their memory capacity and usage. In other words, to obtain the best performance, these models require storing all preceding output masks and the encoded features in their memory, increasing the generalization of the updated model. The resulting memory limitation lead to facing similar challenges already known in the domain of continual learning (below). 

In this paper, we hypothesize that these issues can be mitigated, specifically motivated by the success of continual learning algorithms in preserving the learned knowledge while limiting the required memory.

    \Figure[t](topskip=0pt, botskip=0pt, midskip=0pt)[width=0.85\paperwidth]
	{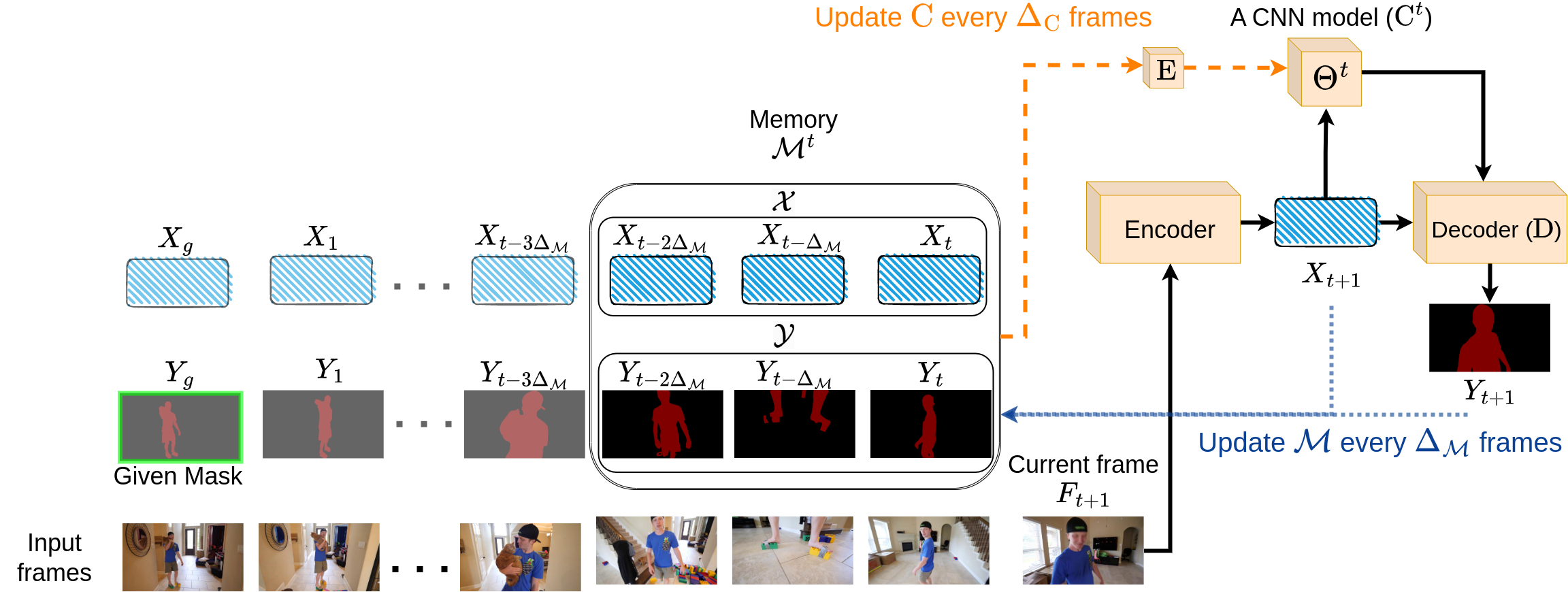}
	{An Online VOS pipeline: The target model $\mathrm{C}^t$ is initialized based on the given ground truth mask \(Y_g\) and its associated feature \(X_g\). The dashed orange line shows how the target model $\mathrm{C}^t$ is updated based on memory $\mathcal{M}^t$ every $\Delta_\mathrm{C}$ frames. The blue dotted arrow illustrates how the memory $\mathcal{M}^t$ is updated every $\Delta_\mathcal{M}$ frames. The methods proposed in this paper are mainly engaged with the target model component of the pipeline.
	\label{fig1}}

\subsection{Continual Learning}

Continual learning~\cite{aljundi2019continual,hsu2018re,li2017learning,yang2021dystab} is a process of sequential learning, where the sequence of data may stem from different domains and tasks; that is, a model is learning from data in which an abrupt or gradual concept drift~\cite{gama2014survey} may take place.

Similarly in Online VOS methods with limited memory, a concept drift can easily happen with regards to the appearance of target objects. In such situations the distribution of the available data in the memory will significantly change with every update step. The primary challenge in this situation is known as {\em catastrophic forgetting}, a term which was first defined in the context of neural networks~\cite{mccloskey1989catastrophic,ratcliff1990connectionist}, although it is a common problem in machine learning~\cite{erdem2005ensemble}.

\subsubsection{Catastrophic Forgetting} Catastrophic forgetting~\cite{kirkpatrick2017overcoming} commonly takes place in machine learning problems such as few shot learning \cite{shi2021overcoming,yap2021addressing}, graph neural networks \cite{liu2021overcoming,zhou2021overcoming} knowledge distillation \cite{binici2022preventing} and Bayesian inference frameworks \cite{chen2021overcoming}.

Catastrophic forgetting occurs when a machine learning model is trained on a sequence of tasks, but at any moment in time it gains access to the training data of only the most recent task. Consequently, the learning has a tendency to update model parameters to be dominated by data from current task. This results in a degree of forgetting previously learned tasks.

In particular, a long video will typically have subsets in which a given object is seen from different view points, varying lighting, different object appearances, occlusion, and missing objects, all of which lead to a continual learning problem. 

For an Online VOS approach each section of a long video in the memory can be considered as a ``task'', thus, forgetting the previously learned tasks (previous parts of the video) can be problematic when processing video sequences, since the number of tasks increases with the length of the video. This problem formulation has been well explained in~\cite{nazemi2023clvos23}, however in this article we focus on developing continual learning-based solutions.

There are three different solutions for catastrophic forgetting problems: prior-focused (regularization-based) \cite{de2021continual,chen2021overcoming}, likelihood-focused (rehearsal-based) \cite{atkinson2021pseudo,castro2018end,wu2019large,zhao2020maintaining}, and hybrid (ensemble) approaches~\cite{lee2016dual,rusu2016progressive}. 

In GPM~\cite{saha2021gradient}, a neural network model takes gradient steps in the opposite direction of the gradient subspace considered relevant for previous tasks in order to learn new tasks. GPM determines the basis of these subspaces by evaluating network representations via a Singular Value Decomposition (SVD) after learning each task. To specify a unique constraint imposed for each layer in a fine-grained fine-tuning regularization, TPGM~\cite{tian2023trainable} proposes an automatic constraint learning method known as the trainable projected gradient method.

In this paper, a regularized (GRCL) solution and a rehearsal-based based (RMSCL) solution are proposed to generalize the the usefulness of Online VOS methods on long video sequences.

\subsection{Feature Selection}
Memory reading is an important step in query-based VOS methods, as they typically employ similarity metrics in their memory reading methods to retrieve and merge partial information from memory for their decoder component. For instance, STCN~\cite{cheng2021rethinking} employs L2 similarity and STM~\cite{oh2019video} utilizes a dot product in their memory reading. This paper aims to enhance Online VOS methods that incorporate online training for a specific component of the model. Therefore, in order to maximize the advantages of memory, it would be beneficial to employ a simple efficient memory selection technique that differs from a complex memory reading. This is because there is no requirement to partially select and merge samples from memory, as is typically done in memory reading methods. In other words, we are searching for memory selection strategies that are covered in feature selection, rather than memory reading in query-based VOS methods.

For data or feature analytics, dealing with high dimensional features greatly increases the need for greater memory and processing power. Additionally, the presence of duplicated, irrelevant, and noisy features arises the likelihood that learning algorithms may loose its generalization ability, which reduces efficiency and performance.

Feature selection methods trying to deal with the high dimensional data are categorized into supervised \cite{thomas2006elements,jian2016multi,li2017challenges} and unsupervised \cite{li2015unsupervised,li2016toward,wei2016unsupervised} learning approaches. 

The discriminative information included in the class labels is accessible to supervised algorithms, but real-world data is typically unlabeled and data annotation is prohibitively costly. Therefore, unsupervised feature selection techniques use several metrics, such as data similarity, density information, and data reconstruction error, to determine the quality of features. 

The Reconstruction based methods approximate the original data by performing a reconstruction function on some selected features \cite{wu2022memory,zhang2020unsupervised,liu2020robust,li2017reconstruction}. In this article we as well propose a Reconstruction-based Memory Selection Continual Learning (RMSCL) to improve Online VOS on long video sequences.

\subsection{Long Video Sequences}
Long video sequences containing several concepts are more challenging to be learned since the model requires a memory with large capacity to store the previously learned frames representations. 

In order to overcome the memory and training time constraints, AFB-URR~\cite{liang2020video} use an exponential moving averages technique to either store a new memory component as it is, or merge it with previous ones if they are related. When the memory's capacity hits a set limit, the model eliminates any features that are not being used.

Using a global context module \cite{li2020fast} is another way to deal with the limitations caused by long video sequences. The model calculates a mean of the entire memory components and apply it as a single representation. 

Nevertheless, the segmentation accuracy is compromised by both approaches as they use a compact representation of the memory. In contrast, XMem~\cite{cheng2022xmem} achieves significantly greater accuracy in both short- and long-term predictions by avoiding compression via the use of a multi-store feature memory. In this article, we focus on improving Online VOS by providing an efficient memory usage method (RMSCL) and a regularization based continual learning approach (GRCL).

\section{Proposed Approach}\label{sec3}
In this section we develop two proposed methods (GRCL and RMSCL) in depth. It is important to understand that the proposed methods apply on Online VOS methods in the evaluation time. Additionally, these methods are not limited to one specific framework, rather they can be extended to any regular Online VOS architecture. The significance of this generality is that Online VOS frameworks are preferred against query-based methods in practical applications, since query-based architectures (such as XMem \cite{cheng2022xmem}) lead to memory requirements which grow with video length, whereas Online VOS methods assumed a fixed memory size. While online learning does not possess the memory challenges associated with query-based methods, online learning-based approaches do have some problems that are addressed in this section.

We begin with the general structure of Online VOS in Section~\ref{sec:olvos}, followed by the formulation of the proposed gated-regularizer (GRCL) in Section~\ref{sec:cl}, and the reconstruction-based memory selection continual learning (RMSCL) in Section~\ref{sec:rms}. We conclude this section by proposing the hybrid method of GRCL and RMSCL.

\subsection{Online VOS}
\label{sec:olvos}
Online VOS~\cite{robinson2020learning,mao2021joint,bhat2020learning}, as overviewed in Figure~\ref{fig1}, typically comprises the following pieces:  
\begin{enumerate}
    \item A pretrained encoder, extracting features from each frame;
    \item A memory $\mathcal{M}^t$, storing features and their associated labels / mask;
    \item A target model $\mathrm{C}^t$, which is trained on the memory at updating time \(t\), and provides information to the decoder;
    \item A label encoder network $\mathrm{E}$~\cite{bhat2020learning} which generates sub-mask labels from each \(Y\) to guide the target model $\mathrm{C}^t$ what to learn from \(Y\).
    \item A Pretrained decoder $\mathrm{D}$ network which obtains temporal information from the target model alongside the encoder's output, to generate a fine-grain output mask $Y_{t+1}$ from the current frame $F_{t+1}$. 
\end{enumerate}
The target model $\mathrm{C}^t$ is usually a small convolutional neural network, for reasons of efficiency.  The target model is updated every $\Delta_\mathrm{C}$ frames throughout the video, repeatedly trained on the complete set of features \(X \in \mathcal{X}\) and the encoded labels $\mathrm{E}(Y)$ of stored decoder outputs \(Y \in \mathcal{Y}\) from  preceding frames. Both \(X\) and \(Y\) are stored in the memory \(\mathcal{M}^t\), where the memory is constrained to maximum size \(N\), as shown in Figure~\ref{fig1} where $N=3$. For online training of $\mathrm{C}^t$, \(Y\) is fed to $\mathrm{E}$ and we seek a trained model $\mathrm{C}^t$ to learn what $\mathrm{E}$ specifies from \(Y\). That is, the target model acts like a dynamic attention model to generate a set of score maps \(\mathrm{E}\big(Y_i\big)\) in order for the target model $\mathrm{C}^t$ to learn and focuses on different important parts of the mask $Y_i$. Thus, $\mathrm{E}$ is only used for training the target model $\mathrm{C}^t$ and is not used  during the inference process. The loss function \(L\) which is used for the online training of target model $\mathrm{C}^t$ is
\begin{align}
\label{eq:frtm}
&L(\Theta^t,\mathcal{M}^t) =  \\  &\sum_{n=1}^{N}\Big\|d_n W_n \Big(\mathrm{E}(Y_n)-\mathrm{C}^{t}(X_n)\Big)\Big\|^2_2 
+\sum_{k=1}^{K}\lambda||{\theta_k^t}^2||, \nonumber 
\end{align}
where $\theta^t_k \in \Theta^t$ is a parameter of $\mathrm{C}^{t}$. Depending on the overall architecture, $\mathrm{E}$ could be an offline / pre-trained label encoder network, as in \cite{bhat2020learning}, or just a pass-through identity function, as in \cite{robinson2020learning}. \(W_n\) is the spatial pixel weight, deduced from \(Y_n\), and $d_n$ is the associated temporal weight decay coefficient. \(W_n\) balances the importance of the target and the background pixels in each frame, whereas \(d_n\) defines the temporal importance of sample \(n\) in the memory, typically emphasizing more recent frames. Here, \(W_n\), \(d_n\), and $\mathrm{E}$ are trained offline and are fixed during the inference time.

Online VOS methods suffer from three main limitations which deteriorate their performance, particularly on long videos:
\begin{enumerate}
\item \textbf{Memory Size:} To maximize performance, Online VOS would need to store in the memory all or most of the extracted information of all preceding frames. However, for videos of arbitrary length this requires an unlimited memory size, which is infeasible.
\item \textbf{Target Model Updating:} Even with an unlimited memory size, updating the target model $\mathrm{C}^t$ on an arbitrarily large memory would be computationally problematic.
\item \textbf{Hyperparameter Sensitivity:} The sensitivity of Online VOS approaches to the target model's configuration and memory updating step size affects both speed and accuracy.
\end{enumerate}
The proposed GRCL and RMSCL aim to mitigate these limitations by incorporating simple yet effective methods applied to the target model $\mathrm{C}^t$ and memory \(\mathcal{M}^t\).

\begin{figure*}[t]
	\begin{center}
		\includegraphics[scale = 0.29]{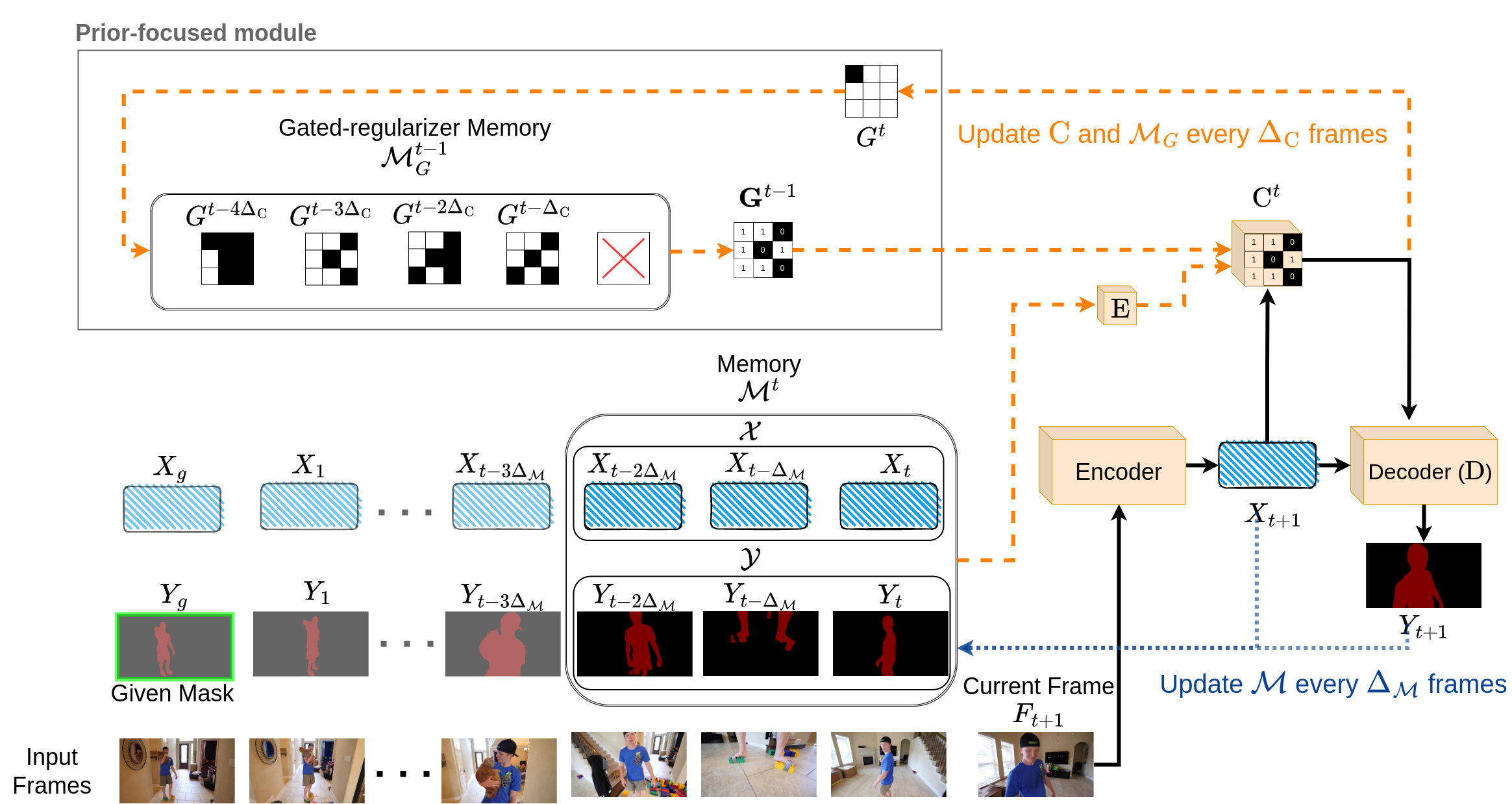}
		
	\end{center}
	\caption{The proposed Online VOS framework, with adopted Gated-Regularized Continual Learning (GRCL): At time \(t\), the overall gated-regularizer map \(\mathbf{G}^{t-1}\) is calculated using the stored gated maps in the gated-regularizer memory \(\mathcal{M}^{t-1}_G\) and regularizes the process of updating $\mathrm{C}^t$. Finally, \(\mathcal{M}^{t-1}_G\) is updated and forms \(\mathcal{M}^{t}_G\) using the calculated \(G^{t}\).}
	\label{fig2}
\end{figure*}

Since video frame information is provided consecutively into the Online VOS framework, there is a high possibility of drift in the object's appearance, especially in long-video sequences. As such, the conventional approach of passing all of the information, as a whole, to the model to decide which to use, is not effective. In the experimental results, we further focus on this specific issue.

Instead, inspired by continual learning~\cite{aljundi2019continual}, we seek to regularize the parameters, \(\Theta^t\), of the target model $\mathrm{C}^t$ in each online learning step \(t\), with a goal of preserving the model knowledge, acquired from those earlier samples (frames) which are no longer present in the memory \(\mathcal{M}^t\).  That is, we have three fundamental questions of

\begin{enumerate}
\item {\em How do we constrain or regularize the model parameters? This question is explored in the gated-regularizer continual learning (GRCL) method of Section~\ref{sec:cl}.
The proposed GRCL is inspired by Memory Aware Synapses (MAS) continual learning~\cite{aljundi2018memory}. The proposed GRCL allows the memory size to be reduced while maintaining model performance, also increasing the robustness of the target model against the updating step size $\Delta_\mathrm{C}$, which otherwise typically affects model  performance. }

\item {\em How do we decide explicitly what to keep in the memory, or which subset of the memory to use in learning? This question is addressed in the context of reconstruction-based memory selection continual learning (RMSCL) of Section~\ref{sec:rms}.
The proposed RMSCL is inspired by reconstruction-based feature selection methods, makes it possible that updating $\mathrm{C}^t$ can efficiently benefit from stored information in the memory \(\mathcal{M}^t\).}

\item {\em What would be the performance of a solution merging RMSCL and GRCL? We name this solution, the Hybrid method, and it is introduced in Section~\ref{sec3d}.
}

\end{enumerate}

\subsection{Regularization-based Continual Learning Solutions}
\label{sec:cl}

Parameter regularization seeks to preserve important parameters of the target model, \(\Theta\), specifically those parameters which were learned or significantly modified in preceding update steps. The MAS algorithm~\cite{aljundi2018memory} is formulated such that at update step $t$ the importance of each parameter \(\theta^t_k\) is associated with its gradient magnitudes \(\{u^l_k\}_{l=1}^{t-1}\) during preceding update steps. Therefore, during each online learning step, we update the parameter weights $\omega^t_k$ based on the gradient magnitudes,
\begin{align}
\label{eq:mas1}
\omega^t_k = \omega^{t-1}_k + u^t_k
\end{align}
To apply a regularization-based continual learning solution such as MAS on Online-VOS with the set of features \(\mathcal{X}\) and their related output masks \(\mathcal{Y}\) in a memory \(\mathcal{M}^t\) having size \(N\), and given a target model $\mathrm{C}^t$ with \(K\) parameters \(\Theta^t\), the regularized loss function \(L_R\) is defined as
\begin{align}
L_R(\Theta^t,\mathcal{M}^t) = L(\Theta^t,\mathcal{M}^t) +
\gamma \sum_{k=1}^{K}\omega^{t-1}_k\big\|\theta^t_k-\theta^{t-1}_k\big\|^2_2,
\label{eq1}
\end{align}
where \(L (\Theta^t, \mathcal{M}^t)\) is as described in~\eqref{eq:frtm}.  The latter term is the regularization, controlled by \(\gamma\), and \(t\) counts the model update steps.

The goal of all regularization-based continual learning solutions such as MAS is that the loss \(L_R\) allows the target model to be updated while preserving its previously important learned knowledge.
Clearly for a method such as MAS, the effectiveness of the loss function \(L_R\) deteriorates over time (frames) as \(\Omega^t = \{\omega_k^t\}_{k=1}^K\) loses its effectiveness in regularization, since most parameters become important as the number of update steps \(t\) increases. This is because MAS only keep a single $\omega^t_k$ for each parameter and accumulate the new calculated parameters importance to the previous one as shown in~\eqref{eq:mas1}. Other memory-based solutions such as Elastic Weight Consolidation (EWC)~\cite{kirkpatrick2017overcoming} keeps the set of gradients magnitude \(\{u^l_k\}_{l=1}^{t-1}\) for each parameter in a memory which is not memory efficient. Our proposed GRCL tries to remove these constraints and broaden the concept to Online VOS.

\subsubsection*{Gated-Regularizer Continual Learning}\label{sec:gb}
We wish to formulate GRCL such that, instead of accumulating the importance parameters in \(\Omega^t\), it stores maximum ($P$) number of binarized importance maps $\{G^j\}_{j=1}^P$ in a dynamic sized gated-regularizer memory \(\mathcal{M}^t_G\) where size of \(\mathcal{M}^t_G\) is limited \(\big(\abs{\mathcal{M}^t_G} \leq P \big)\) and way smaller than the size of the memory $\mathcal{M}^t$.

Thus, at each update step \(t\), the overall gated-regularized map \(\mathbf{G} ^{t-1}\) is defined as
\begin{align}
&\mathbf{G} ^{t-1} = \bigvee_{j =1} ^ J G^{j} \;\;\;, \;\;\; J = \abs{\mathcal{M}^{t-1}_G}
\label{eq:final_gated_regularizer}
\end{align}

Here $ \bigvee$ is the ``Logical 
Or" operator and \(\abs{\mathcal{M}^{t-1}_G}\) is the dynamic size of  $\mathcal{M}^{t-1}_G$. Given the current overall gated-regularizer maps \(\mathbf{G}^{t-1}\), the gated-regularized loss function \(L_G\) can be formulated as
\begin{align}
L_G(\Theta^t,\mathcal{M}^t) = L(\Theta^t,\mathcal{M}^t) +
\gamma \sum_{k=1}^{K}\mathbf{g} ^{t-1}_k\big\|\theta^t_k-\theta^{t-1}_k\big\|^2_2
\label{eq-LG}
\end{align}
where \(\mathbf{g}^{t-1}_k \in \mathbf{G}^{t-1}\), such that with a large coefficient \(\gamma \cong \infty\), it acts as a gating function that allows some parameters to be updated and others to be frozen. After updating the target model $\mathrm{C}^t$, a new gated-map (\(G^{t}\)) should be defined and memory \(\mathcal{M}^{t-1}_G\) is updated.

To this end, after accumulating the magnitude of the gradient in \(U^t = \{u^t_k\}_{k=1}^K\), a binary gated-regularizer \(g^{t}_k \in G^{t}\) will be defined as
\begin{align}
g^{t}_k= \begin{cases} 1
&\text{if~ \(\frac{u^t_k}{\max_{k} (U^t)} > h\)} \\
0 &\text{else} \end{cases}
\label{eq:gated_regularizer}
\end{align}
where \(0<h<1\) is a threshold, which is determined based on the distribution of the gradients in \(U^t\!\). The bigger the value of \(h\), the more sparse the resulting gated-regularized map \(G^{t}\). 

 Figure~\ref{fig2} shows the flow-diagram of an Online VOS framework at time \(t\) when the target model $\mathrm{C}^t$ is regularized by the proposed GRCL.

One of the main advantages in formulating the loss function of the Online VOS framework as \(L_G\) is to store an efficient set of binary maps \(\{G^j\}_{j=1}^P\) in \(\mathcal{M}^t_G\), much smaller in size compared to the sets of features \(\mathcal{X}\) and masks \(\mathcal{Y}\) stored in \(\mathcal{M}^t\) which means $P<<N$.

\subsubsection{Dynamic Gated-Regularizer Memory}
\label{sec:dynamic_memory}

The gated-regularizer memory \(\mathcal{M}_G^t\) can have a fixed size of $P$ similar to \(\mathcal{M}^t\) that has a fixed size of $N$; however, as the number of stored gated-regularized maps is increased, the degrees of freedom of the target model $\mathrm{C}^t$ for learning new information in the memory will be decreased, and that could have negative effects on the performance of the model. To handle this problem, In this section, a dynamic mechanism is proposed to dynamically reduce make the gated-regularizer memory  \(\mathcal{M}_G^t\) of GRCL dynamic in size. To do this, when the overall gated-regularized map \(\mathbf{G} ^{t-1}\) is calculated, the number of ones in \(\mathbf{G} ^{t-1}\) determines the number of regularized parameters of the target model, and if it is smaller than a certain threshold, GRCL tends to expand \(\mathcal{M}_G^t\). On the other hand, if the number of ones in \(\mathbf{G} ^{t-1}\) is greater than another threshold, \(\mathbf{G} ^{t-1}\) will be shrunk, and the oldest stored gated-regularized maps in the memory $\mathcal{M}_G^{t-1}$ will be removed from memory to keep the number of regularized parameters below and above certain thresholds.

The number of regularized parameters upper bound threshold $\eta_u$ is proportionate to the number of target model parameters $K$, thus, the upper-bound of $P$ is when the number of regularized parameters (ones in \(\mathbf{G} ^{t-1}\)) reaches $\eta_u = \xi_u \times K$ and the lower-bound of $P$ is when the number of regularized parameters be less than $\eta_l = \xi_l \times K$.  The two $\xi_u$ and $\xi_l$ ratios would be found for each target model and number of training epochs using cross-validation.

Thus, GRCL does not need to make any changes on Online VOS methods. It only needs to regularize the target model $\mathrm{C}$ updating loss function. Additionally, the hyper-parameters for GRCL that should be tuned are $h$ that is used to binarize the gated maps ($G$) and also two other ratios ($\xi_l$ and $\xi_u$) which determine the the lower bound and upper bound of regularized parameters ($\eta_u$ and $\eta_l$) in $\mathrm{C}$ and make the $\mathcal{M}_G^t$ dynamic in size.

While several other techniques for continual learning have been described in the literature~\cite{abati2020conditional,serra2018overcoming,jung2020continual}, none of them are as well designed for Online VOS as GRCL, with its dynamic gated regularizer memory. It is worth noting that the encoder, decoder and network $\mathrm{E}$ in the proposed architecture are trained offline, and we use the same trained models in all experiments. Additionally, the memory is initialized by the encoded features of the given frame \(F_g\) with the provided ground-truth mask \(Y_g\), as defined in semi-supervised VOS frameworks.

\begin{figure*}[t]
	\begin{center}
		\includegraphics[scale = 0.29]{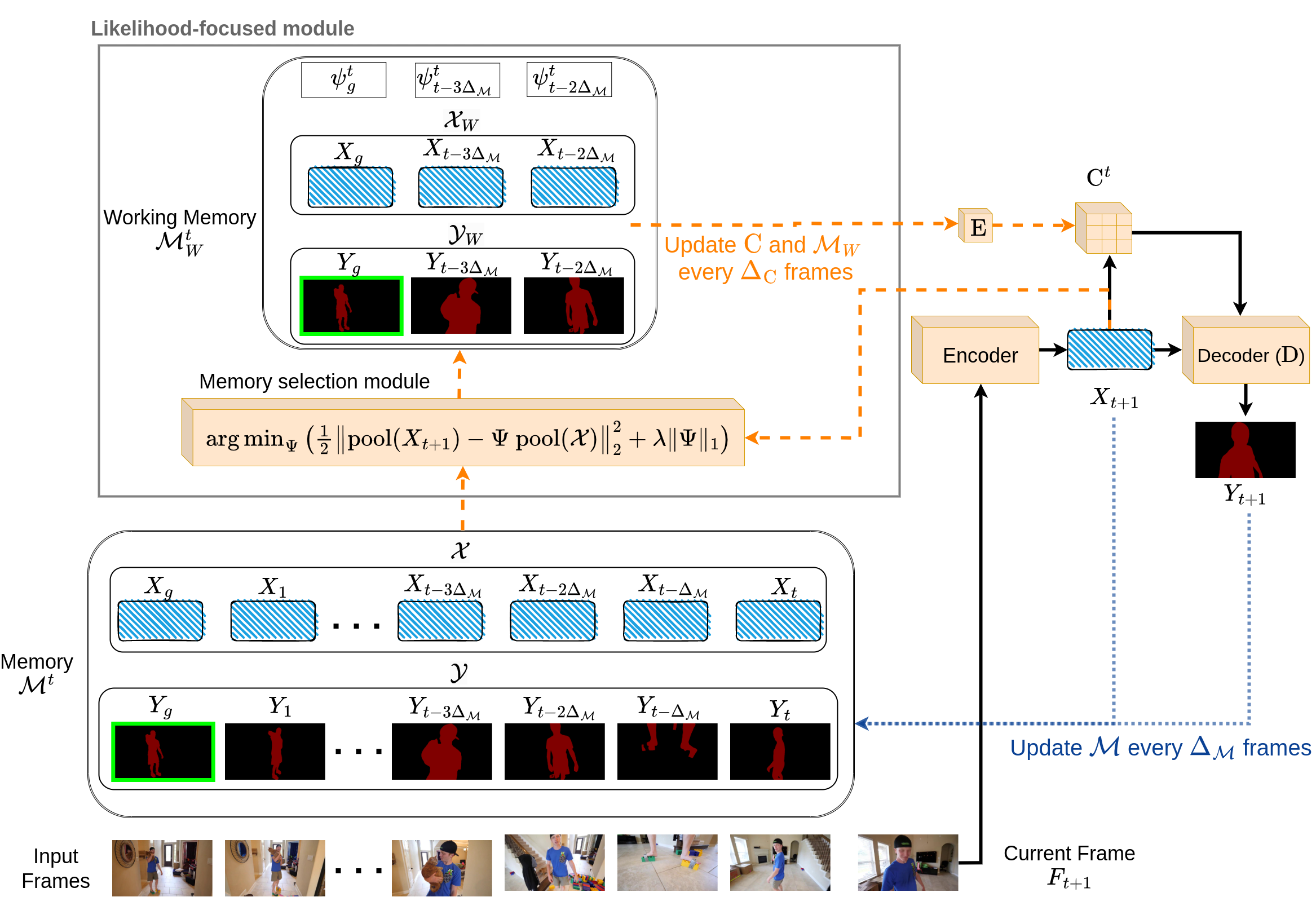}
		
	\end{center}
	\caption{The proposed Online VOS framework with augmented  Reconstruction-based Memory Selection Continual Learning (RMSCL). At the current time \(t\), three samples associated to three positive \(\psi\) are selected using a reconstruction based (Lasso) optimization.}
	\label{fig3-RCS}
\end{figure*}

\subsection{Reconstruction-based Memory Selection Continual Learning}
\label{sec:rms}
Given the forgetting behaviour of an Online VOS due to the appearance drift of objects, a trivial solution for mitigating  this problem is simply to have an unlimited memory size. 
However, it is difficult for a limited-size target model to extract generalized discriminating information from a considerably larger memory \(\mathcal{M}^t\). As such, the effectiveness of updating the target model $\mathrm{C}^t$ becomes dramatically deteriorated on long videos as memory grows.

To solve this limitation, we propose a dynamic working memory \(\mathcal{M}^t_W\), a {\em subset} of \(\mathcal{M}^t\), and update the target model using this new (smaller) memory instead of a (larger) memory \(\mathcal{M}^t\). This new approach can address three problems:
\begin{enumerate}
\item Allowing a limited size target model to benefit from a large memory, and
\item The update step becomes significantly more efficient,  since it is training on a smaller working memory \(\mathcal{M}^t_W\).
\item All of the samples in the memory could have a considerable weight in the training loss function of target model independent of their temporal weight $d_n$, by re-weighting the selected samples from $\mathcal{M}$.
\end{enumerate}
The proposed RMSCL approach adapts a methodology similar to those of likelihood-based (rehearsal) approaches in continual learning, where a set of selected observations from preceding tasks would be preserved in the memory to mitigate the catastrophic forgetting of the target model on proceeding tasks.

As such, \(\mathcal{M}^t_W\) needs to be a small, diverse memory which contains the required features $X$ and masks $Y$ of preceding evaluated frames. Thus, the goal of the proposed RMSCL is to select \(q\) samples from memory \(\mathcal{M}^t\) and to place them in \(\mathcal{M}^t_W\) for target model updating. This memory selection is performed on \(\mathcal{M}^t\) every update step $\Delta_{\mathrm{C}}$ since the goal of creating \(\mathcal{M}^t_W\) is to update the target model $\mathrm{C}$. The selection of samples from memory is formulated as a LASSO~\cite{tibshirani1996regression} optimization problem: To update the target model $\mathrm{C}^{t-1}$, the optimal linear reconstruction of the stored features \(\mathcal{X} \in \mathcal{M}^t\) for the next feature \(X_{t+1}\) is identified via a \(L1\) constraint on a randomly initialized vector of coefficients \(\Psi\) by minimizing
\begin{align}
\label{eq:llc1}
&\Psi^t = \mathop{\arg \min}_{\Psi }L_{RMSCL}(\Psi,\mathcal{M}^t,X_{t+1}) = \\
&\mathop{\arg \min}_{\Psi }\left(\frac{1}{2}\big\| X_{t+1}- \Psi \mathcal{X} \big\| ^ 2_2 + \lambda \|\Psi\|_1\right).\nonumber
\end{align}
It is worth noting that updating the target model $\mathrm{C}^{t-1}$ and creating $\mathrm{C}^{t}$ will happen before segmenting the object in frame $F_{t+1}$ and predicting \(Y_{t+1}\) using the updated $\mathrm{C}^{t}$ at time \(t\). Moreover, $\mathcal{X}$ contains $\abs{\mathcal{M}^t}$ features ($\mathcal{X} = \{X_l\}^{\abs{\mathcal{M}^t}}_{l=1}$), similarly \(\Psi\) consists of $\abs{\mathcal{M}^t}$ coefficients (\(\Psi = \{\psi_l\}_{l=1}^{\abs{\mathcal{M}^t}}\)) weighting each feature \(X_l\) in reconstructing of \(X_{t+1}\). In other words, we want to have the best sparse linear reconstruction of new received frame \(X_{t+1}\) using the stored features $\mathcal{X}$ in memory \(\mathcal{M}^t\).
The \(L_{RMSCL}\) loss leads to a sparse set of coefficients because of $\mathrm{L}_1$-norm in~\eqref{eq:llc1}~\cite{lee2006efficient},  meaning that only a small number of coefficients \(\Psi\) are non-zero after the optimization process, and the positive coefficients \(\psi\) and their associated features \(X\) are selected and are placed in $\mathcal{M}_W^t$ for updating the target model. 
It is important to mention that the deterministic temporal weight \(d_n\) is not involved in $L_{RMSCL}$ loss function in~\eqref{eq:llc1} and instead RMSCL re-weights the selected samples in $\mathcal{M}_W^t$ by the coefficient \(\Psi\) calculated in~\eqref{eq:llc1}. This re-weighting enables RMSCL to include the significance of selected samples in the current update phase. Thus, \(d_n\) is replaced with \(\psi_n\) in~\eqref{eq:frtm} as:
\begin{align}
\label{eq:frtm-rmscl}
&L(\Theta^t,\mathcal{M}_W^t) = \\
&\sum_{n=1}^{\abs{\mathcal{M}_W^t}}\Big\|\psi_n  W_n \Big(\mathrm{E}(Y_n)-\mathrm{C}^{t}(X_n)\Big)\Big\|^2_2 
+\sum_{k=1}^{K}\lambda ~{\theta_k^t}^2.\nonumber
\end{align}
Here, $\abs{\mathcal{M}_W^t}$ is the size of dynamic working memory \(\mathcal{M}_W^t\), equal to number of non-zero positive \(\{\psi_n\}\).
The only problem with the LASSO minimization of~\eqref{eq:llc1} is that its computational complexity depends on the dimensionality of feature \(X\), such that a gigantic feature size can lead optimizing~\eqref{eq:llc1} to becoming the bottleneck of Online VOS. In order to handle this problem, a channel based max pooling function $\text{pool($\cdot$)}$ is applied on each feature \(X\), such that~\eqref{eq:llc1} becomes
\begin{align}
\label{eq:llc2}
&\Psi^t = \mathop{\arg \min}_{\Psi }L_{RMSCL}(\Psi,\mathcal{M}^t,X_{t+1}) \nonumber \\ 
&\simeq \mathop{\arg \min}_{\Psi }\Big(\frac{1}{2}\big\| \text{pool}(X_{t+1})- \Psi \text{pool}(\mathcal{X}) \big\| ^ 2_2 + \lambda \|\Psi\|_1 \nonumber \\
&\text{s.t.} \ \Psi \geq 0 \Big). 
\end{align}
The pooling function pools the feature $X$ with dimension of $C \times W \times H$ to $1 \times W \times H$ by pooling over channels $C$. In other words, the pooling function acts like a dimensionality reduction function which makes the input data be $C$ times smaller in size.
It is worth noting that the pooling function is only performed for estimating the coefficient set \(\Psi\); it is still the actual features \(\mathcal{X}\) which are used for creating the working memory \(\mathcal{M}_W^t\) and updating the target model. Moreover, a constrain is used in~\eqref{eq:llc2} that force $\Psi$ to be not negative.

\subsection{Hybrid method}
\label{sec3d}
Hybrid methods usually benefit from three different continual learning solutions: regularization-based, replay-based, and structural-based~\cite{yoon2017lifelong}. Here, structural-based solutions of continual learning are not used since those models try to expand the model (increasing the parameters of the model) while keeping other important parameters fixed. For an Online VOS solution, expanding the model size over time is not an option since the bottleneck of an Online VOS is the target model, and the computational complexity of Online VOS would be strongly affected by increasing the size of the target model $\mathrm{C}$. The challenges that a Hybrid method of continual learning aims for are better robustness and generalization in different situations. Here, we propose a hybrid approach that takes into account the contributions of both GRCL and RMSCL without considering any other factors. In other words, our proposed Hybrid method will evaluate a solution that has both working memory and gated-regularizer memory in its structure. 
The loss function $L_H$ that is used for the proposed Hybrid solution is
\begin{align}
&L_H(\Theta^t,\mathcal{M}_W^t,\mathbf{G}^{t-1}) = \\
& L(\Theta^t,\mathcal{M}_W^t) +
\gamma \sum_{k=1}^{K}\mathbf{g} ^{t-1}_k\big\|\theta^t_k-\theta^{t-1}_k\big\|^2_2. \nonumber
\label{Hybrid_Loss}
\end{align}

The evaluation of the proposed methods are explained in the next section.

\section{Results}\label{sec4}
The effectiveness of the proposed methods to improve the performance of Online VOS frameworks is evaluated by augmenting state-of-the-art Online VOS algorithms. It is worth noting that both proposed gated-regularizer continual learning (GRCL) and reconstruction-based memory selection continual learning  (RMSCL) can augment a given Online VOS framework, and they can even be combined and used together as the ``Hybrid'' approach.

Here we adopt two well-known and state-of-the-art Online VOS frameworks:  LWL~\cite{bhat2020learning} and JOINT~\cite{mao2021joint}.
LWL is an extension over the well-known FRTM~\cite{robinson2020learning} framework, benefiting from a label encoder network $\mathrm{E}$ which tells the target model what to learn~\cite{bhat2020learning}. JOINT approaches the VOS problem by using an online learning induction branch, jointly with a transduction branch which benefits from a lightweight transformer for providing sufficient temporal and spatial attention to its decoder. JOINT has reported the state-of-the-art performance for the problem of Online VOS in terms of accuracy.

\subsection{Datasets}
We compared the proposed methods with two different types of video sequences: long and short. The long video dataset~\cite{liang2020video} contains objects with a long trajectory with multiple distribution drifts; the short videos are from the standard  DAVIS16~\cite{perazzi2016benchmark}, DAVIS17~\cite{perazzi2016benchmark}, and YouTube-VOS18~\cite{xu2018youtube} datasets, where the target objects are being tracked in a short period of time and usually without significant changes in appearance.
Evaluating the competing methods on both long and short video datasets demonstrates the robustness of different algorithms to different environments. 

The \textbf{Long Video Dataset}
~\cite{liang2020video} contains three videos with a single object which is recorded for more than \(7000\) frames. The target objects could have some sudden appearance changes, which lead to significant representation drifts of the video objects. Each of the videos in the dataset has $21$ labelled frames for the evaluation.

With regards to \textbf{Short Video Datasets},
the DAVIS16~\cite{perazzi2016benchmark} validation set has \(20\) videos, each of which has a single object for segmentation, the validation set of DAVIS17~\cite{perazzi2016benchmark} contains \(30\) video sequences with multiple objects to be segmented in each frame. The validation set of YouTube-VOS18 has \(474\) video sequences of \(65\) seen (which are present in the training set) and \(26\) unseen object classes. The target objects in these datasets are mostly with a short trajectory, with modest changes in object appearance.

\subsection{Experimental Setup}

We use a fixed parameter setup for the baselines, with maximum memory sizes of \(N=32\) for LWL and  \(N=20\) for JOINT, as is suggested in their setups. For all experiments, the target model $\mathrm{C}^t$ is updated for three epochs in each updating step to have a fair comparison with baselines. The target model is updated every time the memory is updated, following the proposed setup in~\cite{cheng2022xmem}. The memory \(\mathcal{M}^t\) is initialized based on the given (ground truth) frame \(F_g\).

In all experiments, as suggested in the semi-supervised Online VOS baselines (LWL and JOINT), the information in \(F_g\) is preserved and is used throughout the whole video sample. For GRCL, we keep the gated-regularizer map \(G^0\) related to the training of $\mathrm{C}^0$ in \(\mathcal{M}^t_G\). For RMSCL, the feature \(X_g\)  and mask $Y_g$ are always placed in working memory with a minimum weight \(\psi_g\) as shown in Figure~\ref{fig3-RCS}.
We use the same available pre-trained decoder and encoder models for all experiments of LWL and JOINT. 
To measure the effectiveness of the competing methods, consistent with the standard DAVIS protocol~\cite{perazzi2016benchmark}, the mean Jaccard \(\mathcal{J}\) index, mean boundary \(\mathcal{F}\) scores, and the average of \(\mathcal{J}\& \mathcal{F}\) are reported for all of the methods. For YouTube-VOS18, the reported results are found using the YouTube-VOS18 official evaluation server~\cite{xu2018youtube}. The mean Jaccard \(\mathcal{J}\) index, mean boundary \(\mathcal{F}\) scores, and the overall score \(\mathcal{G}\) of seen and unseen object classes are also reported. The speed of each method is reported on the DAVIS16 dataset~\cite{robinson2020learning} in units of Frames Per Second (FPS) when $\Delta_{\mathrm{C}} = \Delta_{\mathcal{M}} = 1$.
All experiments were performed using a single NVIDIA V100 GPU.

\begin{figure}[t]
	\begin{center}
		\includegraphics[scale = 0.72]{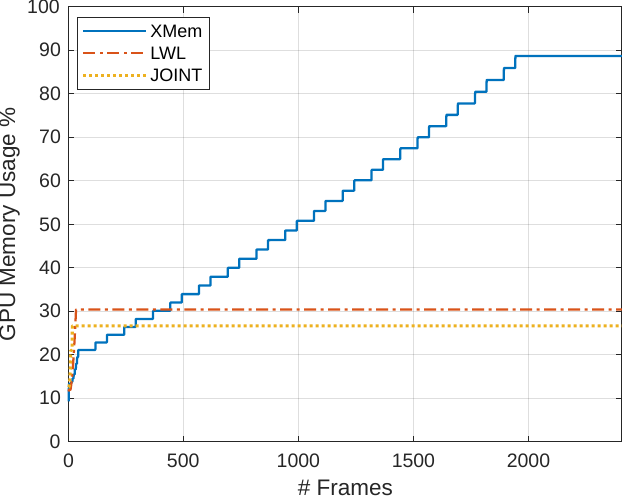}
		
	\end{center}
	\caption{ GPU memory usage of XMem, LWL and JOINT when processing $2416$ frames of the \(\boldsymbol{blueboy}\) video in the long video dataset~\cite{liang2020video}. As shown, the GPU memory usage of XMem increases significantly over time, whereas LWL and JOINT have a fixed GPU memory usage.}
	\label{fig0-xmem-LWL-JOINT}
\end{figure}

\begin{table}
\centering
\caption[Comparative results on the Long Videos dataset~\cite{liang2020video}]{Comparison results on the Long Videos dataset~\cite{liang2020video}, based on the Online VOS baseline methods (LWL and JOINT), their augmented versions with GRCL, RMSCL, Hybrid, and four matching-based VOS methods. The evaluation metric $\mathcal{J}$ is related to the Intersection over Union (IoU) of an estimated object mask and the ground truth, and $\mathcal{F}$ is about how accurate is the boundary.}
\vspace{10 pt}
\begin{tabular}{l|ccc}
\hline
    \bf Method    & \bf \(\mathcal{J\&F}\) &\bf \(\mathcal{J}\) &\bf \(\mathcal{F}\)  \\

\hline
            LWL & 79.8 \(\pm\) 4.2 & 78.0 \(\pm\) 4.3 & 81.6 \(\pm\) 4.2
            \\

            LWL-GRCL (ours) &   84.5 \(\pm\) 1.6  & 82.8 \(\pm\) 1.3 & 86.1 \(\pm\) 2.0
            \\
            LWL-RMSCL (ours) &   83.4 \(\pm\) 2.7 & 81.5 \(\pm\) 2.6 & 85.2 \(\pm\) 2.8            
            \\
            LWL-Hybrid (ours) &   \textbf{85.4 \(\pm\) 1.0} & 84.0 \(\pm\) 1.0 & 86.9 \(\pm\) 1.1
            \\
            \hline
			JOINT & 67.5 \(\pm\) 4.4 & 65.7 \(\pm\) 4.2 & 69.3 \(\pm\) 4.7
			\\
			JOINT-GRCL  (ours)& 70.5 \(\pm\) 6.8 & 68.7 \(\pm\) 6.6 & 72.3 \(\pm\) 7.0
			\\
   			JOINT-RMSCL (ours)& \textbf{75.6 \(\pm\) 5.1} & 73.6 \(\pm\) 5.0 & 77.5 \(\pm\) 5.2
			\\
      		JOINT-Hybrid (ours)& 72.3 \(\pm\) 5.3 & 70.4  \(\pm\) 5.2 & 74.0 \(\pm\) 5.3
			\\
            \hline
			RMNet & 59.8 \(\pm\) 3.9 & 59.7 \(\pm\) 8.3 & 60.0 \(\pm\) 7.5
			\\       
   			STM & 80.6 \(\pm\) 1.3 & 79.9 \(\pm\) 0.9 & 81.3 \(\pm\) 1.0
			\\    
   			STCN & 87.3 \(\pm\) 0.7 & 85.4 \(\pm\) 1.1 & 89.2 \(\pm\) 1.1
			\\    
   			XMem & \textbf{89.8 \(\pm\) 0.2} & 88.0 \(\pm\) 0.2 & 91.6 \(\pm\) 0.2
			\\    
\bottomrule
\end{tabular}
\label{tbl1}
\end{table}

\begin{figure*}[h]

\begin{center}

\begin{tabular}{cc}
\includegraphics[scale = 0.82]{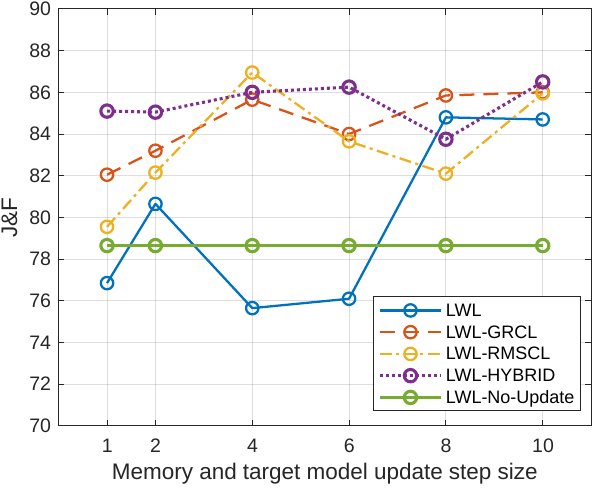}&
\includegraphics[scale = 0.85]{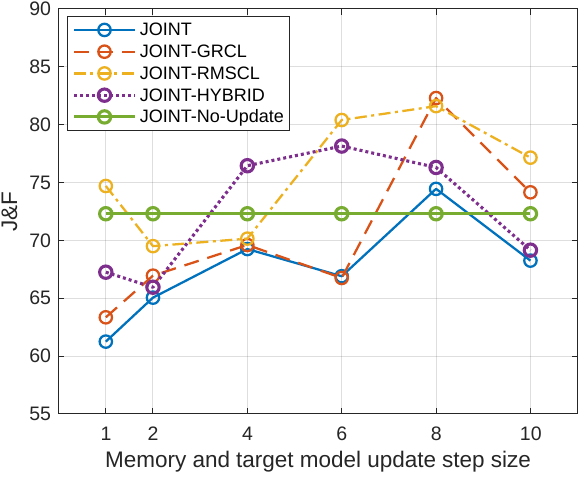}
\end{tabular}
\caption
{Performance comparison of competing methods as a function of memory and target model update step sizes, (\(\Delta_{\mathrm{C}}=\Delta_{\mathcal{M}}\)), on the Long Videos dataset~\cite{liang2020video}. The left figure shows the average \(\mathcal{J}\& \mathcal{F}\) of applying different proposed methods on LWL and the right one shows the performance of the same methods on JOINT. The green line shows the performance of LWL and JOINT without updating their target model on the memory.}
\label{fig:four-rows-results}
\end{center}
\end{figure*}

\begin{figure*}
	\begin{center}
		\includegraphics[scale = 0.59]{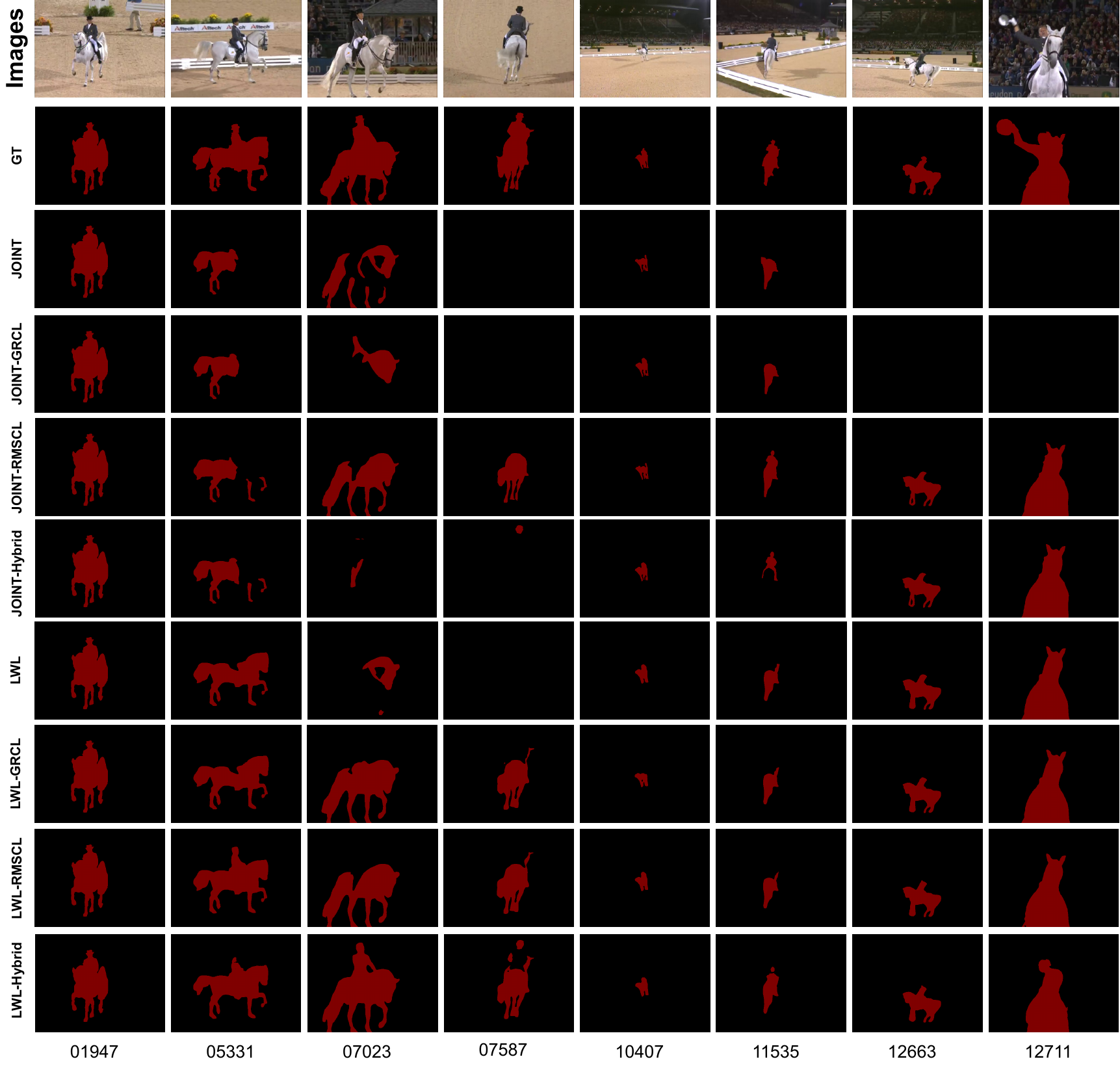}
		
	\end{center}
	\caption{Qualitative comparison of the competing frameworks in the context of the long-video dataset. The associated frame number for each image is shown along the bottom. The leftmost column shows the given mask \(Y_g\), which is the same for all methods. The results show that the proposed GRCL, when augmenting the baseline frameworks (LWL and JOINT), can lead to better performance against representation drift. Additionally, the frameworks based on RMSCL (LWL-RMSCL, JOINT-RMCSL) are less vulnerable to the distribution changes which take place in long video sequences. Finally, as shown in the figure, LWL-Hybrid has the best performance among all the proposed methods.}
	\label{fig-qual1}
\end{figure*}

\begin{table*}[t]
\centering
\caption [Performance of comparative method on three short video datasets.]{Performance analysis of the evaluated methods against the validation sets of DAVIS16, DAVIS17, and YT-VOS18.}
\begin{tabular}{l|ccccc|ccc|ccc|c}
\toprule
    \multirow{2}{*}{Method}  & \multicolumn{5}{c|}{YT-VOS 2018} & \multicolumn{3}{c|}{DAVIS17} & \multicolumn{3}{c|}{DAVIS16}& \multirow{2}{*}{FPS}
    \\ 
    \cmidrule{2-12}
  & \(\mathcal{J}_s\) & \(\mathcal{F}_s\) &\(\mathcal{J}_u\) & \(\mathcal{F}_u\) & \(\mathcal{G}\)&
  
  \(\mathcal{J}\) & \(\mathcal{F}\) & \(\mathcal{J}\&\mathcal{F}\) &
  
  \(\mathcal{J}\) & \(\mathcal{F}\) & \(\mathcal{J}\&\mathcal{F}\)& \\ 
\midrule
			LWL~\cite{bhat2020learning} & 79.5& 83.9& 75.7 & 83.4 & 80.6&  77.1 & 82.9 &80.0& 87.3 & 88.5 &87.9& 18.87
			\\
   			LWL-GRCL (ours)&79.7& 84.0&75.9 & 83.9& 80.9&  77.0 & 83.0 & 80.0  & 87.3 & 88.6 &88.0 & 15.89
			\\
      		LWL-RMSCL (ours)& 79.4& 84.0& 75.2 & 83.2 & 80.5 & 75.6 & 81.8 & 78.7 & 86.5 & 88.3 & 87.4& 19.01
			\\
            LWL-Hybrid (ours)& 77.2& 81.4& 71.8 & 79.9& 77.6& 72.7& 79.2 & 76.0& 83.8 & 85.5 & 84.7 & 16.19
            \\
   \hline
   			JOINT~\cite{mao2021joint} & 81.6& 85.6& 78.6 & 86.0& 82.9&  80.6 & 86.0 &83.3& 87.5 & 89.4 &88.5& 6.21
			\\
   			JOINT-GRCL (ours)& 81.6& 85.5& 77.6 & 84.9 & 82.4& 80.4 & 85.8 &83.2& 87.5 & 89.4 &88.5& 6.09
			\\
      		JOINT-RMSCL (ours)& 81.0& 85.0& 77.6 & 84.9 & 82.2& 79.8 &85.4&82.6& 87.9& 90.0 & 89.0& 11.15
			\\
            JOINT-Hybrid (ours)& 81.6&85.6& 78.6 & 85.9 & 82.9&79.9 & 85.4& 82.6 &87.8  & 89.9 & 88.9 & 10.8
            \\
   \hline
			RMNet~\cite{xie2021efficient} & 82.1& 85.7& 75.7 & 82.4& 81.5& 81.0 & 86.0 &83.5& 88.9 & 88.7 &88.8& 11.9
			\\
   			STM~\cite{oh2019video} & 79.7& 84.2& 72.8 & 80.9& 79.4& 79.2 &84.3 &81.8& 88.7 & 89.9 &89.3& 14.0
			\\
   			STCN~\cite{cheng2021rethinking} & 81.8& 86.5& 77.9 & 85.7& 83.0& 82.2 & 88.6 &85.4& 90.8 & 92.5 &91.6& 26.9
			\\
   			XMem~\cite{cheng2022xmem} & 84.6& 89.3& 80.2 & 88.7& 85.7& 82.9 & 89.5 &86.2& 90.4 & 92.7 &91.5& 29.6
			\\
\bottomrule
\end{tabular}
\label{tbl2}
\end{table*}

\subsection{Experimental Results}
\label{sec:results-43}
\subsubsection{Long Video Evaluation}
Figure~\ref{fig0-xmem-LWL-JOINT} shows the GPU memory usage of LWL, JOINT and XMem on the ``blueboy'' video sequence from the long video dataset. Online VOS methods (LWL and JOINT) require only a fixed GPU memory size, which enables them to be used on smaller devices with more modest GPUs. This section will show that the proposed methods do not further increase the GPU memory requirement while they improve the performance of Online VOS methods.

The effectiveness of the proposed GRCL and RMSCL is evaluated by augmenting two state-of-the-art Online VOS frameworks, LWL and JOINT, however our proposed methods can be extended to any Online VOS method having a periodically-updated target model network, as in Figure~\ref{fig1}. In addition to the proposed GRCL and RMSCL methods, we show the proposed Hybrid method which is a combination of GRCL and RMSCL could improve the performance of GRCL and RMSCL in some cases.

Table~\ref{tbl1} shows the results of the selected baselines (LWL and JOINT), each augmented by the proposed GRCL, RMSCL and Hybrid, evaluated on the Long video dataset.
For LWL-GRCL and JOINT-GRCL, the threshold \(h\) is dynamically set to the \({99.5}^{th}\) percentile of the distribution of normalized \(U^t\) in~\eqref{eq:gated_regularizer}. Additionally, $h$ is limited \((0.1<h<0.55)\) for LWL-GRCL and \((0.1<h<0.6)\) for JOINT-GRCL. Bounding the threshold $h$ prevents the model from regularizing many not important parameters or very few important parameters. The hyper-parameters related to \(h\) were selected by cross-validation. 

The chosen ratios of GRCL ($\xi_l$ and $\xi_u$) are $0.07$ and $0.15$, respectively. These ratios are defined for the target model $\mathrm{C}$ and are identical for LWL and JOINT. Taking these two ratios into account, the upper and lower bounds of number regularized parameters for LWL and JOINT are $\eta_u = 0.15 \times K$ and  $\eta_u = \xi_u \times K$, respectively. For instance, for LWL with a target model with $K=73728$ parameters (weights), two chosen upper and lower limit thresholds ($\eta_l$ and $\eta_u$) on the number of regularized parameters of the target model $\mathrm{C}$ would be $5000$ and $11000$, respectively. 

For the adopted frameworks by RMSCL, the parameter \(\lambda\) defines the sparsity of \(\Psi\) in~\eqref{eq:llc2}. To select the best \(\lambda\), Akaike Information Criterion (AIC)~\cite{akaike1974new,aho2014model} is used for model selection,  automatically selecting \(\lambda\) and the number of positive non-zero coefficients \(\Psi^t\), which defines the size of the working memory \(\mathcal{M}_W^t\). Thus, for each update step $t$, in principle \(\mathcal{M}_W^t\) could have a different size in comparison to $\mathcal{M}^t$, depending upon the selected \(\lambda\), the current feature $X_{t+1}$, and the set features \(\mathcal{X}\) in the memory $\mathcal{M}^t$.

It is worth noting that the selected hyper-parameters for Hybrid solution are the sames as the selected parameters for GRCL and RMSCL for each dataset assuming the Hybrid method benefits from the best version of both GRCL and RMSCL and does not need to re-tune new hyper-parameters for its GRCL and RMSCL parts.

We conduct six experiments with six different memory and target model update steps \(s \in \{1,2,4,6,8,10\}\), where the target model $\mathrm{C}^t$ is updated after each memory update. The performance of RMSCL  fluctuates with update step size $\Delta_\mathrm{C}$, because of the differing distributions which are formed in the memory as a function of sampling frequency. For reference,  the means and standard deviations of all competing methods are reported in Table~\ref{tbl1}. 

In \cite{cheng2022xmem}, authors also compare the performance of different methods by taking the average of five runs, however, they did not report the five update steps which they used. Comparing the standard deviations of JOINT in Table~\ref{tbl1} with those reported in \cite{cheng2022xmem}, we see that our six selected memory update steps are close to those in~\cite{cheng2022xmem}.

As seen in Table~\ref{tbl1}, the proposed methods improve the performance of both Online VOS models on long videos when the objects in the video have a long trajectory with sudden representation drifts. LWL-Hybrid method improves the average performance \(\mathcal{J\&F}\) of LWL by more than \(4 \%\), while JOINT-RMSCL improves JOINT more than \(8 \%\).

Furthermore, as illustrated in Table~\ref{tbl1}, the proposed methods improve the robustness of LWL model against different memory \(\mathcal{M}^t\) and model $\mathrm{C}$ update step sizes evident by the lower reported standard deviation in the Table~\ref{tbl1}. 
It is worth mentioning that JOINT has a parallel transduction branch in its structure, which benefits from a transformer model that acts like a matching-based method. 
Although the transduction branch of JOINT can boost the positive or even negative effects of the proposed solutions, the average performance \(\mathcal{J\&F}\) of JOINT is improved significantly.

As shown in Table~\ref{tbl1}, using a Hybrid method improves the robustness (smaller standard deviation) of the baselines with a better average performance on LWL. However, on JOINT, JOINT-Hybrid does not have better performance than JOINT-RMSCL and it shows a better Hybrid method is needed for JOINT. It is worth mentioning that JOINT-Hybrid has better performance than JOINT.
To have a fair comparison, the proposed methods and the baseline Online VOS frameworks are compared with four matching-based methods including RMNet~\cite{xie2021efficient}, STM~\cite{oh2019video}, STCN~\cite{cheng2021rethinking}, and the current long VOS state-of-the-art approach XMem~\cite{cheng2022xmem}. The reported results of the matching-based methods on short video datasets are from~\cite{cheng2022xmem}.
STM is a query-based VOS baseline which had been state-of-the-art method for a long period of time in VOS and RMNet, STCN and XMem are its follow up methods. RMNet and STCN try to improve the memory functionality of STM by having a better memory encoding and memory reading methods. XMem also can be considered as an extension of STM which is specifically designed to work on long video sequences.

\begin{figure*}[t]
	\begin{center}
		\includegraphics[scale = 0.39]{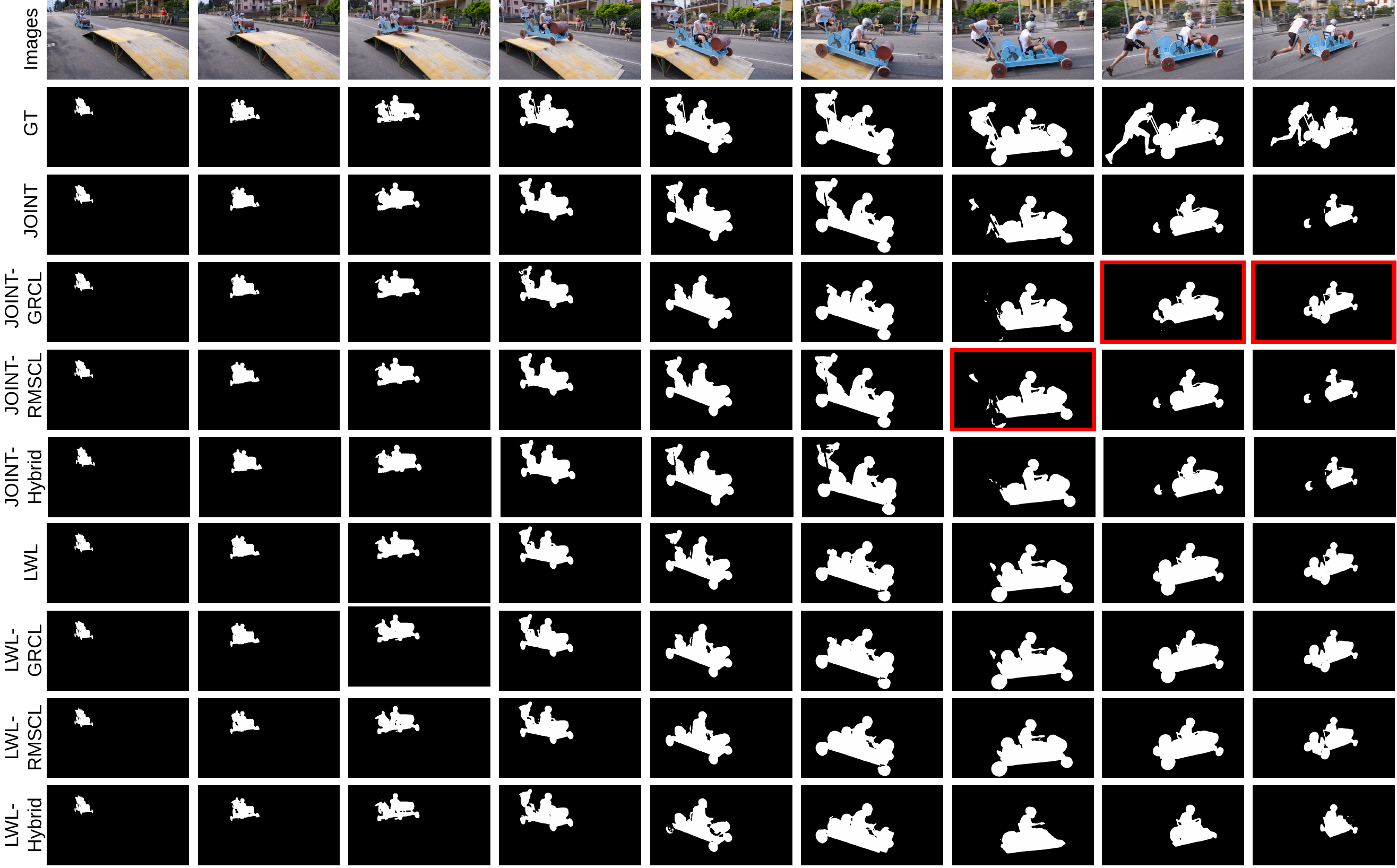}
		
	\end{center}
	\caption{The qualitative comparison of the evaluated methods on a short video dataset (DAVIS16~\cite{perazzi2016benchmark}). The results show that the proposed GRCL and RMSCL can improve the performance of JOINT on DAVIS16 while not having much negative effects on the LWL framework. These qualitative results reflects the quantitative results of Table~\ref{tbl2}.}
	\label{fig-qual2}
\end{figure*}

\begin{figure}[t]
	\begin{center}
		\includegraphics[scale = 0.72]{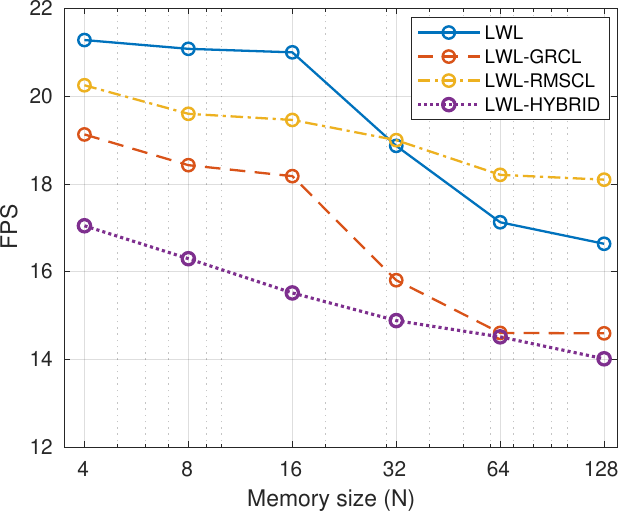}
		
	\end{center}
	\caption{Run-time evaluation:  The proposed methods' run-times are compared against the LWL baseline. LWL-RMSCL reported higher Frame Per Second (FPS) when the memory size \(N\) is increased.}
	\label{fig3-memory-fps}
\end{figure}

As demonstrated in Figure~\ref{fig:four-rows-results}, the average performance \(\mathcal{J\&F}\)
of each \(6\) runs based on different memory and target model update step sizes $\Delta_{\mathrm{C}}$ are compared. In other words, Figure~\ref{fig:four-rows-results} shows first eight methods performance of Table~\ref{tbl1}. On LWL, GRCL outperforms RMSCL in most cases when (\(\Delta_\mathrm{C} = \Delta_{\mathcal{M}} = \{1,2,6,8,10\}\)) whereas on JOINT,  RMSCL is better than GRCL and it is because of the effect of working memory on both branches of JOINT. It is worth noting that GRCL can only impact the induction branch (online learning part) of JOINT.

Figure~\ref{fig-qual1} shows the qualitative results of the proposed methods (GRCL, RMSCL, and Hybrid) and baselines (LWL and JOINT) on \(7\) selected frames of the ``dressage'' video sequence from the Long Videos dataset. The results in Figure~\ref{fig-qual1} are produced by applying the evaluated methods to the Long Videos dataset when $\Delta_{\mathrm{C}} = \Delta_{\mathcal{M}} = 1$. The results show RMSCL improves the performance of LWL and JOINT; however, GRCL improves the performance of LWL but cannot improve the performance of JOINT on the selected frames. Thus, GRCL improves the performance of the baselines if the prior information (information in the memories) is correct (it happens with LWL). As shown in the figure, LWL-Hybrid has the best performance on the Long Videos dataset; however, baseline methods are more vulnerable to the distribution drift of the target object. The distribution drifts that happen on long videos such as the ``dressage'' video are particularly explained, discussed and formulated in~\cite{nazemi2023clvos23}.

\subsubsection{Short Video Evaluation }
Table~\ref{tbl2} demonstrates the performance of adopted Online VOS frameworks based on the proposed approaches and competing algorithms on short video datasets (i.e., DAVIS16, DAVIS17, and YouTube-VOS18). The same hyper-parameters are used for short and long videos, meaning that the models do not have prior knowledge of the length of the video sequence being processed. Objects in short video datasets have a short trajectory and their representations are mostly kept intact or gradually changing through the frames. As seen in Table~\ref{tbl2}, the augmented frameworks by the proposed GRCL perform the same as the baseline methods, and the proposed regularizer not only does not affect the performance of the baseline method when there is no representation drift on objects in videos, but also LWL-GRCL performs slightly better compared to LWL on YouTube-VOS18.

In Table~\ref{tbl2}, we follow the baseline models' suggested parameters for reporting \(\mathcal{J}\), \(\mathcal{F}\) and FPS. For JOINT, \(\mathcal{M}^t\) is updated every \(3\) frames, and for LWL \(\mathcal{M}^t\) is updated every frame; however, XMem update its so called working memory every \(5\) frames. 
The proposed RMSCL improves the performance of JOINT on DAVIS16 but it slightly degrades the performance of JOINT on DAVIS17 and YouTube-VOS18. In JOINT-RMSCL both its online learning part and its transformer part use $\mathcal{M}^t_W$ and that is why JOINT-RMSCL reports higher FPS in comparison with JOINT. RMSCL and consequently the Hybrid method degrade the performance of LWL on DAVIS17 and DAVIS16 showing that memory selection method can be improved for short videos.
Table~\ref{tbl2} also shows the baselines perform slightly better than GRCL in terms of FPS since GRCL needs to calculate a new \(G^{t}\) after every updating step \(t\); however, for a small target model $\mathrm{C}^t$ this FPS degradation is not significant.

We compare the qualitative results of proposed methods and baselines on a short video dataset (DAVIS16~\cite{perazzi2016benchmark}). Figure~\ref{fig-qual2} shows the qualitative results of evaluated models on the ``soapbox'' video sequence of DAVIS16. As illustrated, the proposed methods offer positive improvement on JOINT results  with slight and tiny changes of LWL results, which is in agreement with the reported results in Table \ref{tbl2}. 
 ``Soapbox'' video sample is considered as one of the longest video sequences of DAVIS16 with \(99\) frames.

\begin{figure}[t]
	\begin{center}
		\includegraphics[scale = 0.72]{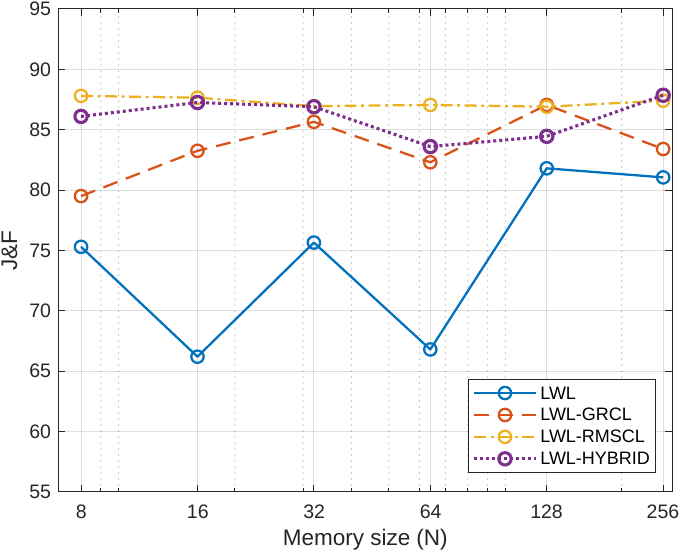}
		
	\end{center}
	\caption{The effect of different memory size \(N\) against the proposed methods compared to the baselines on the long video dataset~\cite{liang2020video}. The performance of the LWL baseline fluctuates with memory size (\(N\)). Here, the target model and memory update step are \(\Delta_{\mathcal{M} } = \Delta_{\mathrm{C}} = 4\).}
	\label{fig3-memory}
\end{figure}

 On long video sequences, it is not feasible to store all of the previously evaluated frames' information in the memory \(\mathcal{M}\), as such, it is important to limit the memory size \(N\). Here, we aim to evaluate how different memory sizes affect baselines and the proposed methods. For this experiment, we compare the performance of LWL, LWL-RMSCL, LWL-GRCL, and LWL-Hybrid on the Long Videos dataset \(N \in \{8,16,32,64,128\}\) and the target model and memory update step are \(\Delta_{\mathcal{M} } = \Delta_{\mathrm{C}} = 4\). As seen in Figure~\ref{fig3-memory}, increasing the memory size \(N\) improves the performance of the methods, but it also increases the computational complexity of the evaluated Online VOS methods. Increasing the size of memory $\mathcal{M}$ does not have a considerable effect on LWL-RMSCL since it does not have any hyper-parameters that are affected by the size of $\mathcal{M}$; however, tuning the hyper-parameters ($P, h, \xi_l, \xi_u$) of LWL-GRCL and consequently LWL-Hybrid is implicitly affected by the size of the memory $N$. This is the reason that LWL-GRCL and LWL-Hybrid performance fluctuated with changing the size of $\mathcal{M}$. It is worth noting that the size of $\mathcal{M}$ RMSCL on the other hand, provides a small set of diverse enough data with new weights $\Psi$ in its dynamic working memory \(\mathcal{M}_W^t\) which improves both accuracy and speed of the baseline methods on long video datasets.
\begin{figure}[t]
	\includegraphics[scale = 0.72]{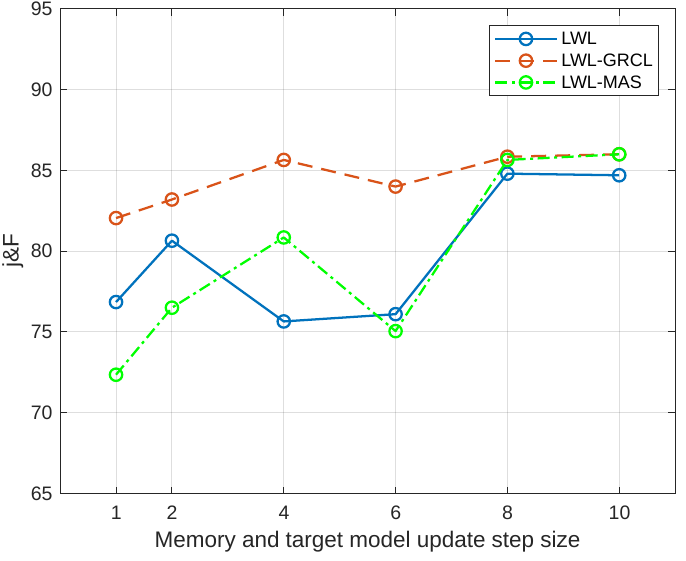}
	\caption{ Quantitative evaluation of the proposed GRCL with conventional continual learning  (MAS)~\cite{aljundi2018memory} on the long video dataset~\cite{liang2020video}. As seen, standard MAS~\cite{aljundi2018memory} is not  as effective as the proposed GRCL when incorporated into the Online VOS framework.}
	\label{fig5-mas-comparison}
\end{figure}

Figure~\ref{fig3-memory-fps} illustrates how increasing the memory size \(N\) affects the speed, measured in FPS, of the evaluated methods (LWL, LWL-GRCL, LWL-RMSCL, and LWL-Hybrid) on DAVIS16. The memory and the target model update step is set to $\Delta_{\mathcal{M}} = \Delta_{\mathrm{C}} = 1$ for the results in Figure~\ref{fig3-memory-fps}. As shown, the FPS of LWL-RMSCL is degraded less than LWL, LWL-GRCL and LWL-Hybrid while LWL-GRCL and LWL reported almost the same degradation with increasing the memory size \(N\). Since LWL-RMSCL uses smaller working memory $\mathcal{M}_W^t$ for training the target model, it would be faster than LWL and LWL-GRCL when the memory size is increased (bigger than $32$). It is worth mentioning that minimizing~\eqref{eq:llc2} in the RMSCL approach is affected by increasing the memory size \(N\) and consequently it affects the FPS of LWL-RMSCL as well. LWL-Hybrid has the lowest FPS among the proposed approaches since it has the computational complexity of both GRCL and RMSCL. When the memory size ($N$) is increased, the Hybrid approach degrades at the same rate as RMSCL.

\subsubsection{Conventional Continual Learning}
One important aspect of the proposed continual learning methods to augment Online VOS frameworks is that they are customized and designed specially for this purpose. 
To illustrate that, here we compare  the performance of the proposed methods (LWL-GRCL) against when the LWL framework is augmented by a standard MAS continual learning modules~\cite{aljundi2018memory} as a regularizer for updating the target model. The evaluation is conducted on long video dataset where the results are demonstrated in Figure~\ref{fig5-mas-comparison}.
As shown in Figure~\ref{fig5-mas-comparison}, LWL-GRCL reported higher average performance \(\mathcal{J\&F}\) compared to when LWL is augmented with MAS. Two main reasons can be provided to further elaborate the reported gap on the performance of these two compared frameworks: i) The overall gated-regularized map \(\mathbf{G}^{t-1}\) described in Figure~\ref{fig3-RCS} keeps the efficiency of the proposed GRCL compared to the MAS approach, where the MAS regularizer loses its efficiency when the update steps are increased. MAS highly benefits from $\Omega^t$, however the efficiency of $\Omega^t$ is being degraded as more and more target model gradients are processed, accumulated, and stored over time. It causes all of the parameters to become important as the number of updates increases. On the other hand, LWL-GRCL with its dynamic memory size, guarantees that the target model $\mathrm{C}^t$ has enough free parameters to learn new tasks. ii) For a small number of training epochs, in each updating step of $\mathrm{C}^t$ the binarized regularizer $\mathbf{G}^{t-1}$ (hard regularizer) is more effective than MAS with a soft regularizer $\Omega^t$.

\subsubsection{Memory Efficiency}
To compare the memory efficiency of the proposed GRCL against the baseline, we compare each unit of memory $\mathcal{M}$ of LWL and each memory unit of $\mathcal{M}$ of adopted LWL-GRCL.
In LWL, each sample in the memory $\mathcal{M}$ consists of the preceding estimated object masks \(\mathcal{Y}\) and its related input frames' extracted features \(\mathcal{X}\). Each feature \(X \in \mathcal{X}\) has a dimension of $512\times30\times52$ floats ($64$ bits). In contrast, each binary regularized-gated map ($G$) has a dimension of $512\times16\times3\times3$ bits. Moreover, each unit of $\mathcal{M}$ also has a binary mask of the target model $\mathrm{C}$ output size of $30\times52$. As a result, each unit of $\mathcal{M}_G^t$ is almost $693$ times smaller than each unit of $\mathcal{M}$. This comparison would be more important if there was a need to improve the performance of an Online VOS implemented on a small device. Thus, having a larger gated-regularizer memory $\mathcal{M}_G^t$ is less expensive than having a large memory $\mathcal{M}$.

\begin{figure}[t]
	\begin{center}
		\includegraphics[scale = 0.8]{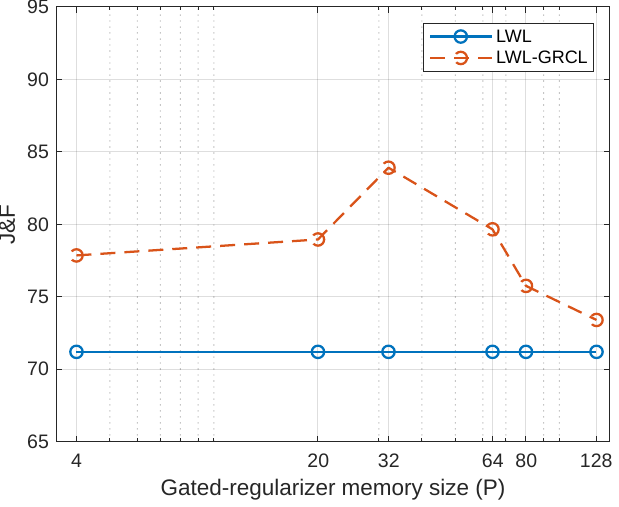}
		
	\end{center}
	\caption{The effect of regularized-gated memory size \(P\) on the LWL-GRCL framework with a fixed size gated memory $\mathcal{M}_G^t$. For this experiment, the memory size $\mathcal{M}^t$ is fixed (\(N = 8\)) in order to properly analyze the impact of the proposed GRCL. By setting \(P\) to a large number, the target model $\mathrm{C}^t$ will not have enough free parameters to be updated on memory \(\mathcal{M}^t\).}
	\label{fig4-gated-memory}
\end{figure}

\begin{figure}[t]
	\begin{center}
		\includegraphics[scale = 0.72]{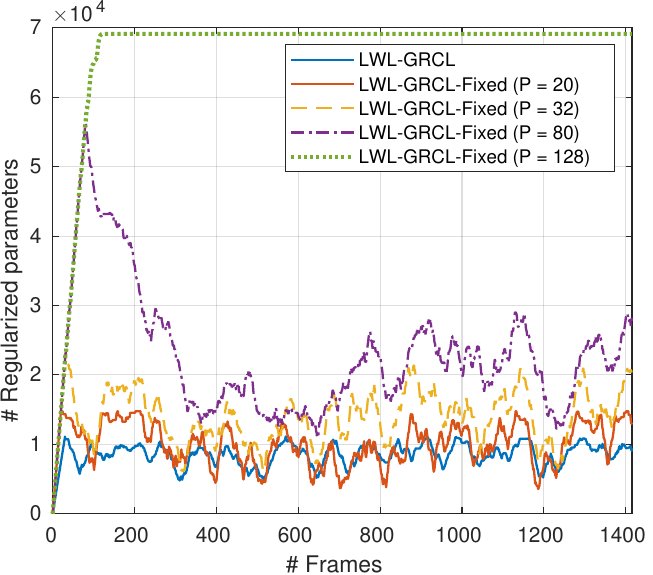}
		
	\end{center}
	\caption{The number of regularized target model parameters when incorporated into LWL-GRCL, as a function of regularized-gated memory sizes (\(P = \{20,32,80,128\}\)). The result are based on $1416$ frames of the \(\boldsymbol{rat}\) video sequence of the long video dataset~\cite{liang2020video}. For this experiment, $\mathrm{C}^t$ is updated every frame and the memory size is set to eight (\(N = 8\)).}
	\label{fig-target modelodel capacity}
\end{figure}

\subsection{Ablation Study}
In this section, we evaluate and analyze the effect of some key parameters of the proposed methods on the performance of both LWL and JOINT methods when augmented with GRCL, RMSCL and Hybrid. For the experimental results, the gated-regularizer memory \(\mathcal{M}^t_G\) has a dynamic size and we control this size by two parameters ($\xi_l, \xi_u$) using cross validation. However, if we consider different fixed memory sizes $P$ for both LWL-GRCl and JOINT-GRCL we could get different results and the goal is to get a reasonable good result with smallest possible regularized parameters set. Here, we evaluate the effect of different gated-regularizer memory size \(P\) of LWL-GRCL on long video dataset. 
\subsubsection{Gated-memory Size}
\label{sec:subsub-gated-reg}
Figure~\ref{fig4-gated-memory} shows the performance of LWL-GRCL with different gated-memory size \(P \in \{4,20,32,64,80,128\}\). It is worth noting that in Figure~\ref{fig4-gated-memory}, LWL-GRCL has a fixed gated memory size. As demonstrated in Figure~\ref{fig4-gated-memory}, increasing \(P\) improves the performance of LWL-GRCL till the number of regularized parameters do not degrade target model learning. The best value for  $P$ depends on $N$. For instance, when the memory size is $N = 8$, LWL-GRCL with a fixed $P = 32$ has the best performance.

\subsubsection{Regularized Parameters}
The number of regularized parameters in $\mathrm{C}^t$ is important factor related to the ability of the target model to learn new information.
As seen in Figure~\ref{fig-target modelodel capacity}, the regularized parameters of the target model $\mathrm{C}^t$ is increased, while the gated-regularizer memory \(\mathcal{M}_G\) is growing by evaluating new frames (the evaluation of first $P$ frames). This growth in the number of regularized parameters is different for different $P$ for GRCL-Fixed where the gated-regularizer memory size is fixed. For instance, for \(P=128\), almost all of the parameters of $\mathrm{C}$ are regularized, and in this case $\mathrm{C}^t$ does not have any free parameters to be trained, and even removing or replacing one gated-regularization map \(G^j\) from $\mathcal{M}_G$ would not free enough parameters and solve the problem. In other words, $\mathrm{C}^t$ would not have enough free parameters to be updated on the new updated memory $\mathcal{M}^t$. When the gated-regularizer memory $\mathcal{M}_G$ reaches its maximum capacity, the oldest $G$ in the gated-regularizer memory will be replaced by the next gated-regularizer map $G^t$. This will free up some parameters since the current $G^t$ usually has a lower number of ``1" in its map in comparison with the oldest $G$ in the memory. This decreasing number of regularized parameters in the overall gated-regularizer $\mathbf{G}^{t-1}$ will continue until it reaches the balance number, and then this phenomenon will continue in a periodic order. This can be seen more clear in Figure~\ref{fig-target modelodel capacity} on LWL-GRCL-Fixed (P=80). The same pattern with smaller intensity happens to the other two, LWL-GRCL-Fixed (P=32) and LWL-GRCL-Fixed (P=20). To address the discussed issue in GRCL-Fixed, a mechanism is proposed for GRCL that makes $\mathcal{M}_G$ dynamic in size. For LWL-GRCL, no $P$ is set, and as shown in Figure~\ref{fig-target modelodel capacity} and as you can see, the number of the target model's regularized parameters is bounded between $5000$ and $11000$.

\subsubsection{Target Model Update Step Size $\Delta_{\mathrm{C}}$}
\begin{figure}[t]

\begin{center}

\includegraphics[scale = 0.69]{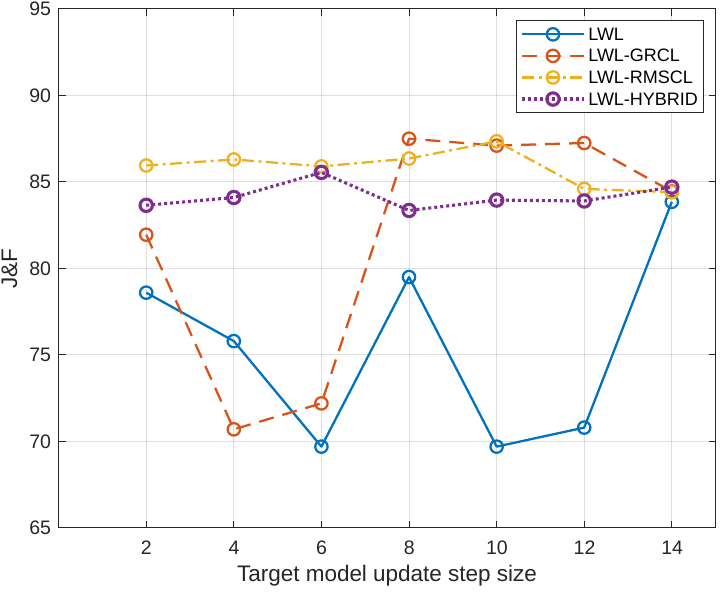}

\caption{The effect of target model update step size $\Delta_{\mathrm{C}}$:  The competing methods are evaluated via long video dataset~\cite{liang2020video}. The results show that the proposed LWL-RMSCL is more robust to target model update step size $\Delta_\mathrm{C}$.}
\label{fig:update-size}
\end{center}
\end{figure}

To demonstrate the effect of target model update step size on the proposed approaches, an ablation study on the Long Videos dataset compares the performance of LWL-GRCL, LWL-RMSCL, LWL-Hybrid, and LWL. The memory update step size is set to \(\Delta_{\mathcal{M}}=1\), whereas the target model update step size varies \(\Delta_{\mathrm{C}} \in \{2,4,6,8,10,12,14\}\). The memory $\mathcal{M}$ and $\mathrm{C}^t$ were updated sequentially at the same time index ($\Delta_{\mathrm{C}} = \Delta_{\mathcal{M}}$). For this experiment, the memory capacity is set to \(N = 4\), making the situation difficult for all of the evaluated approaches. Figure~\ref{fig:update-size} shows that LWL has the lowest performance when compared to LWL-GRCL and LWL-RMSCL. LWL's update step size $\Delta_{\mathrm{C}} \in \{6,10,12\}$ decreases, however the proposed methods reduce the degree of performance degradation, except for LWL-GRCL at $\Delta_{\mathrm{C}}=4$. This decrease of performance by GRCL demonstrates a limitation of the GRCL method when the model concentrates on an incorrect prior, which is maintained and amplified during the evaluation of future frames. 
\section{Conclusion}\label{sec12}
In this paper, we  proposed two novel modules, Gated-Regularizer Continual Learning (GRCL) and Reconstruction-based Memory
Selection Continual Learning (RMSCL), which can be integrated with any Online VOS algorithms and improve their memory-limitation while improving their performance on long videos and preserving their performance accuracy on short videos. Another important benefits of the proposed approaches is that the offline trained parts of the adopted methods (LWL and JOINT) which are the encoder, decoder $\mathrm{D}$, and the label encoder $\mathrm{E}$, do not need to be re-trained for each proposed method. This makes it possible to apply the proposed methods on any Online VOS that follows the explained general Online VOS in Figure~\ref{fig1} without any fine-tuning process.

Additionally, we showed the combination of two proposed method (proposed Hybrid method) will increase the performance of the augmented baselines usually when GRCL and RMSCL are in their best performance. It is important to mention that our results showed a need for designing a better Hybrid method which can have a smarter combination of GRCL and RMSCL. The proposed methods improve the performance of Online VOS in different scenarios and aspects such as speed, accuracy and robustness. Moreover, the proposed regularization-based method (GRCL) does not mitigate the performance of the baselines on the evaluated short video datasets (DAVIS16, DAVIS17, and YouTube-VOS18). 

\bibliographystyle{unsrt}
\bibliography{sn-bibliography}

\begin{IEEEbiography}[{\includegraphics[width=1in,height=1.25in,clip,keepaspectratio]{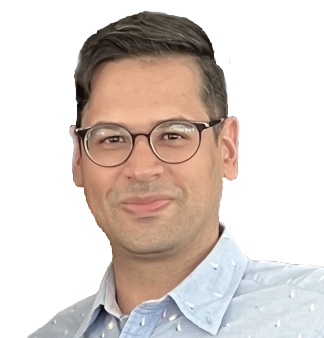}}]{Amir Nazemi} currently works as a postdoctoral researcher at the Department of Systems Design Engineering at the University of Waterloo. He has worked on many research projects with ETRI, Microsoft, and the Faculty of Health at the University of Waterloo. He received his B.Sc. and M.Sc. degrees in Computer Software Engineering and Artificial Intelligence in 2010 and 2014, respectively, and his Ph.D. in systems design engineering with a title ``Continual learning-based Video Object Segmentation'' from the University of Waterloo in Canada in 2023. His primary research interests are in continual learning and video object segmentation on long videos, computer vision, machine learning, specifically generative AI, and medical imaging.

\end{IEEEbiography}

\begin{IEEEbiography}[{\includegraphics[width=1in,height=1.25in,clip,keepaspectratio]{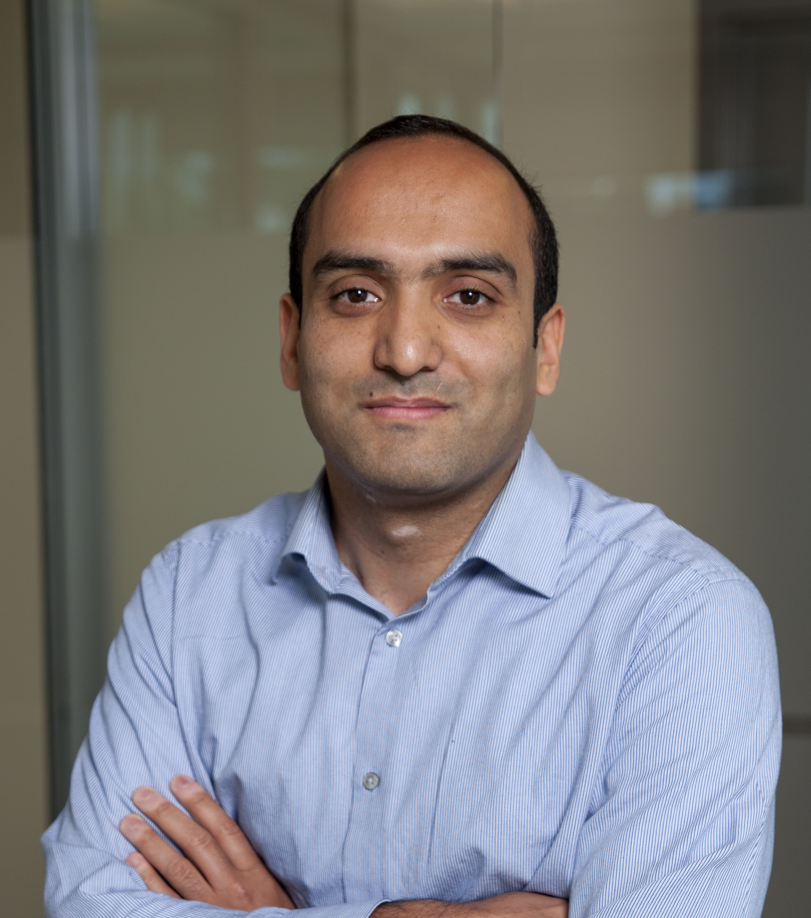}}]{Mohammad Javad Shafiee} is currently the Cofounder and VP Research at DarwinAI and an Adjunct Professor at the Department of Systems Design Engineering at University of Waterloo. He received the B.Sc. and M.Sc. degrees in Computer Science and Artificial Intelligence, in 2008 and 2011 respectively; and the Ph.D. degree in systems design engineering with the focus on Machine Learning and Deep Learning from the University of Waterloo, Canada in 2017. His main research
focus is on statistical learning and graphical models with random fields and deep learning approaches. His research interests include Computer Vision, Machine Learning and Biomedical Image Processing.
\end{IEEEbiography}
\begin{IEEEbiography}[{\includegraphics[width=1in,height=1.25in,clip,keepaspectratio]{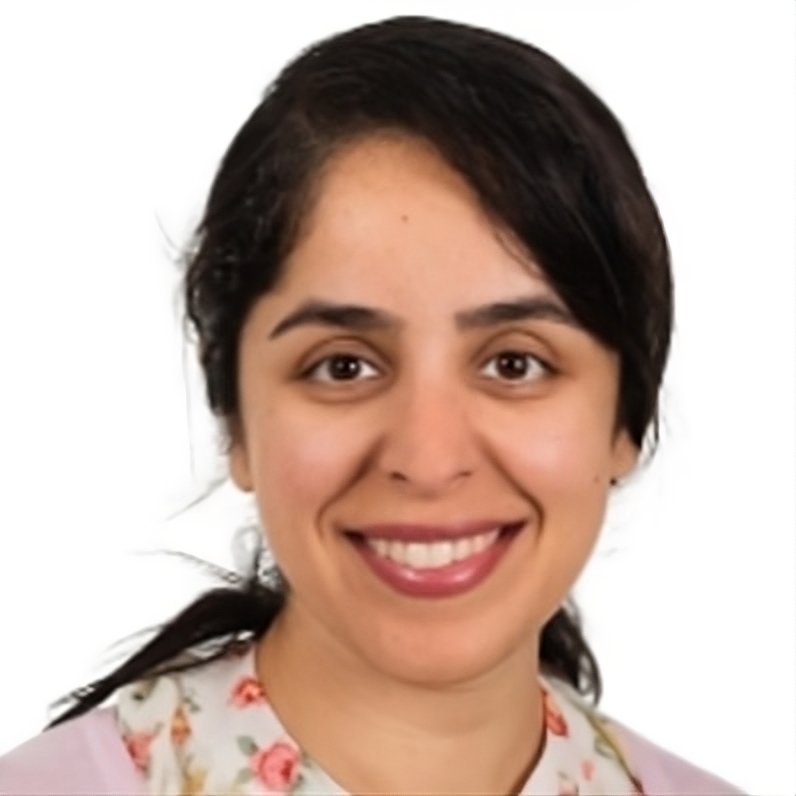}}]{Zahra Gharaee} received the B.Sc. degree in electrical control engineering and the M.Sc. degree
in mechatronics from K. N. Toosi University of Technology, Iran, in 2009 and 2012, respectively, and the Ph.D. degree in cognitive science from the Department of Philosophy and Cognitive Science, Lund University, Sweden, in 2018. She
was a Postdoctoral Researcher with the Computer Vision Laboratory (CVL), Department of Electrical Engineering, Linköping University, Sweden, from 2018 to 2022. She is currently a Postdoctoral Research Fellow with the Vision and Image Processing Laboratory (VIP), Department of Systems Design Engineering, University of Waterloo, Canada. She has been involved
in different research projects with WASP, EU-FET WYSIWYD, Microsoft, and BIOSCAN. Her research interests include artificial intelligence, machine learning, and computational cognitive science.
\end{IEEEbiography}

\begin{IEEEbiography}[{\includegraphics[width=1in,height=1.25in,clip,keepaspectratio]{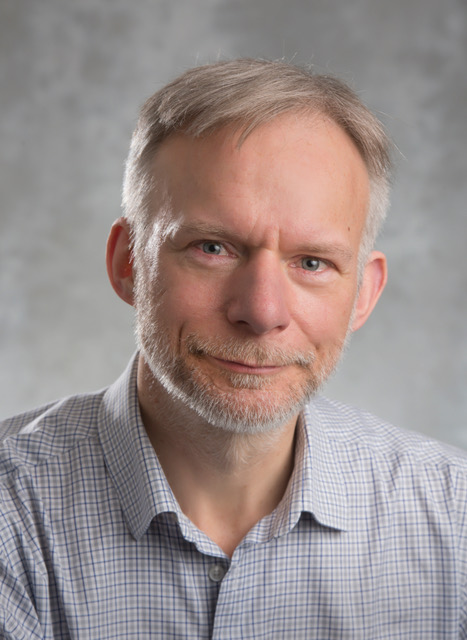}}]{Paul Fieguth} studied undergraduate Electrical Engineering at the University of Waterloo and graduate engineering degrees at the Massachusetts Institute of Technology (MIT).  He has been a member of the faculty at the University of Waterloo in Systems Design Engineering since 1996, where he has been Associate Chair Undergraduate, Department Chair, Associate Dean and, since 2023, Associate Vice President.
His research interests include statistical signal and image processing, hierarchical algorithms, data fusion, machine learning, and the interdisciplinary applications of such methods.  He has significant pedagogical interests in the area of complex systems, specifically developing a much deeper understanding among engineering students on the impact of complex systems in many areas of engineering decision making.  He is the author three textbooks, a 2010 text on Statistical Image Processing \& Multidimensional Modeling, a 2021 text on Complex Systems, and a 2022 text on Pattern Recognition.

\end{IEEEbiography}





\EOD

\end{document}